\documentclass[fleqn]{article} 

\usepackage{jfrStyle}
\usepackage{graphicx}
\graphicspath{{figures/}}
\usepackage{subfig}
\usepackage{apalike}
\let\bibhang\relax
\usepackage{natbib}
\usepackage{setspace}
\usepackage{csvsimple}
\usepackage{lineno}
\usepackage{cite}
\usepackage{url}
\usepackage{hyperref}
\usepackage{siunitx}
\usepackage[margin=1in,footskip=.5in]{geometry}
\usepackage{setspace}
\usepackage{verbatim}
\usepackage{textgreek}
\usepackage[utf8]{inputenc}
\usepackage{amssymb}
\usepackage{color}
\usepackage{amsmath}
\usepackage[toc]{appendix}
\usepackage{lscape}

\usepackage{xcolor}
\usepackage[most]{tcolorbox}
\newtcolorbox{highlighted}{colback=white,breakable}
\newcommand\hl[1]{%
  \bgroup
  \hskip0pt\color{blue!90!black}%
  #1%
  \egroup
}

\usepackage{lmodern}
\usepackage{mathtools, nccmath}

\usepackage{xparse}
\DeclarePairedDelimiterX{\set}[1]{\{}{\}}{\setargs{#1}}
\NewDocumentCommand{\setargs}{>{\SplitArgument{1}{;}}m}
{\setargsaux#1}
\NewDocumentCommand{\setargsaux}{mm}
{\IfNoValueTF{#2}{#1} {#1\,\delimsize|\,\mathopen{}#2}}%{#1\:;\:#2}
\usepackage{nomencl}
\makenomenclature

\usepackage{etoolbox}
\renewcommand\nomgroup[1]{%
  \item[\bfseries
  \ifstrequal{#1}{S}{Symbols}{%
  \ifstrequal{#1}{A}{Abbreviations}{%
  \ifstrequal{#1}{C}{Other Symbols}{}}}%
]}

\usepackage{titlesec}
\titleformat{\paragraph}
{\normalfont\normalsize\bfseries}{\theparagraph}{1em}{}
\titlespacing*{\paragraph}
{0pt}{3.25ex plus 1ex minus .2ex}{1.5ex plus .2ex}

\usepackage[ruled,vlined]{algorithm2e}
\usepackage{array,tabularx,calc}

\title{Active learning with MaskAL reduces annotation effort for training Mask R-CNN}

\author{
Pieter M. Blok \\
Wageningen University \& Research\\
Wageningen, The Netherlands \\
\texttt{pieter.blok@wur.nl} \\
\And
Gert Kootstra \\
Wageningen University \& Research\\
Wageningen, The Netherlands \\
\texttt{gert.kootstra@wur.nl} \\
\And
Hakim Elchaoui Elghor \\
Exxact Robotics\\
Épernay, France \\
\texttt{hakim.elchaoui@exxact-robotics.com} \\
\And
Boubacar Diallo \\
Exxact Robotics\\
Épernay, France \\
\texttt{boubacar.diallo@exxact-robotics.com} \\
\AND
Frits K. van Evert \\
Wageningen University \& Research\\
Wageningen, The Netherlands \\
\texttt{frits.vanevert@wur.nl} \\
\And
Eldert J. van Henten \\
Wageningen University \& Research\\
Wageningen, The Netherlands \\
\texttt{eldert.vanhenten@wur.nl} \\
}

\providecommand{\keywords}[1]
{
  \textbf{\textit{Keywords:}} #1
}

\begin{document}
% \Cref{algo1}
% \lipsum[1-5]

% \linenumbers
% \doublespacing

\maketitle

\begin{abstract}
The generalisation performance of a convolutional neural network (CNN) is influenced by the quantity, quality, and variety of the training images. Training images must be annotated, and this is time consuming and expensive. The goal of our work was to reduce the number of annotated images needed to train a CNN while maintaining its performance. We hypothesised that the performance of a CNN can be improved faster by ensuring that the set of training images contains a large fraction of hard-to-classify images. The objective of our study was to test this hypothesis with an active learning method that can automatically select the hard-to-classify images. We developed an active learning method for Mask Region-based CNN (Mask R-CNN) and named this method MaskAL. MaskAL involved the iterative training of Mask R-CNN, after which the trained model was used to select a set of unlabelled images about which the model was most uncertain. The selected images were then annotated and used to retrain Mask R-CNN, and this was repeated for a number of sampling iterations. In our study, MaskAL was compared to a random sampling method on a broccoli dataset with five visually similar classes. MaskAL performed significantly better than the random sampling. In addition, MaskAL had the same performance after sampling 900 images as the random sampling had after 2300 images. Compared to a Mask R-CNN model that was trained on the entire training set (14,000 images), MaskAL achieved 93.9\% of that model's performance with 17.9\% of its training data. The random sampling achieved 81.9\% of that model's performance with 16.4\% of its training data. We conclude that by using MaskAL, the annotation effort can be reduced for training Mask R-CNN on a broccoli dataset with visually similar classes. Our software is available on \url{https://github.com/pieterblok/maskal}.
\end{abstract}


%\mbox{}
\clearpage
\nomenclature[A]{CNN}{convolutional neural network}
\nomenclature[A]{mAP}{mean average precision}
\nomenclature[A]{Mask R-CNN}{mask region-based convolutional neural network}
\nomenclature[A]{MaskAL}{active learning software for Mask R-CNN}
\nomenclature[A]{RGB}{red, green and blue}
\nomenclature[A]{NMS}{non-maximum suppression}
\nomenclature[A]{GNSS}{global navigation satellite system}
\nomenclature[A]{COCO}{common objects in context}
\nomenclature[A]{PAL}{probabilistic active learning}
\nomenclature[A]{DO}{dropout}
\nomenclature[A]{RPN}{region proposal network}
\nomenclature[A]{FC}{fully connected}
\nomenclature[A]{ANOVA}{analysis of variance}
\nomenclature[A]{ROI}{region of interest}
\nomenclature[A]{USA}{United States of America}
\nomenclature[A]{UK}{United Kingdom}
\nomenclature[A]{NL}{Netherlands}
\nomenclature[A]{AUS}{Australia}
\nomenclature[A]{GNSS}{global navigation satellite system}
\nomenclature[S]{$M_1$}{first mask}
\nomenclature[S]{$M_2$}{second mask}
\nomenclature[S]{$\tau_\texttt{IoU}$}{threshold on the intersection over union}
\nomenclature[S]{$\cap$}{intersection}
\nomenclature[S]{$\cup$}{union}
\nomenclature[S]{$S$}{instance set}
\nomenclature[S]{$s$}{instance belonging to an instance set}
\nomenclature[S]{$fp$}{number of forward passes}
\nomenclature[S]{$K$}{set of available classes}
\nomenclature[S]{$k$}{class}
\nomenclature[S]{$H$}{entropy value}
\nomenclature[S]{$H_\texttt{max}$}{maximum entropy value}
\nomenclature[S]{$H_\texttt{sem}$}{entropy value of an instance}
\nomenclature[S]{$c_\texttt{sem}$}{semantic certainty}
\nomenclature[S]{$c_\texttt{box}$}{spatial certainty of the bounding box}
\nomenclature[S]{$c_\texttt{mask}$}{spatial certainty of the mask}
\nomenclature[S]{$c_\texttt{spl}$}{spatial certainty}
\nomenclature[S]{$B$}{bounding box}
\nomenclature[S]{$\bar{B}$}{average bounding box}
\nomenclature[S]{$M$}{mask}
\nomenclature[S]{$\bar{M}$}{average mask}
\nomenclature[S]{$c_\texttt{occ}$}{occurrence certainty}
\nomenclature[S]{$c_\texttt{h}$}{instance certainty}
\nomenclature[S]{$c_\texttt{avg}$}{average certainty of all instances in the image}
\nomenclature[S]{$c_\texttt{min}$}{minimum certainty of all instances in the image}
\nomenclature[S]{$P$}{Mask R-CNN's confidence score on a class}
\nomenclature[S]{$I$}{instance sets in an image}
\nomenclature[S]{$\texttt{c}_{\texttt{h}_{fp}}$}{certainty value at a specific forward pass}
\nomenclature[S]{$\texttt{c}_{\texttt{h}_{100}}$}{certainty value at 100 forward passes}
\nomenclature[S]{$n$}{the number of classes}
\nomenclature[S]{$r$}{the number of instances belonging to an instance set}
\nomenclature[S]{$t$}{the number of instance sets in an image}
\printnomenclature[3 cm] 

{\clearpage}
\section{Introduction}
\label{introduction}
In current practice, broccoli heads are harvested by hand, and this is physically demanding, time consuming, and expensive. These labour problems can be mitigated by a robot that can harvest the broccoli heads automatically. For the robot to be autonomous, it is essential to have a perception system that can determine which broccoli heads are both healthy and large enough to be harvested. This perception system can be realised with a camera and an image processing algorithm.

Much research has been done on the image-based detection and size estimation of broccoli heads \citep{ramirez2006,kusumam2017,bender2020,lelouedec2020,montes2020,blok2021, blok2021_sizing, garciamanso2021, psiroukis2022}. Unfortunately, the methods presented in previous studies were not able to detect individual diseases and defects in the broccoli crop (\citet{garciamanso2021} did investigate broccoli disease detection, but clustered all diseases and defects as one class "wasted"). Individual detection of diseases and defects is desirable, as this would allow the broccoli harvesting robot to perform specific treatments for each broccoli disease and defect. This disease treatment functionality can increase the economic viability of the robot.

With the current state-of-the-art convolutional neural networks (CNNs), it is possible to learn the broccoli diseases and defects as separate classes. One of the challenges for optimising the CNN, is the selection and annotation of a sufficient number of representative images. Image selection can be challenging when diseases and defects occur only sporadically in the field and thus in the images. Image annotation can be challenging for multiple reasons. First, it can be difficult to correctly label the diseases and defects as they can be visually similar. Second, the annotation process might require additional input from crop experts with specific knowledge about the disease, and this can make the annotation process more time consuming and expensive. Third, it is also desirable to annotate the pixels of each broccoli head, as this enables another algorithm to estimate the size of the broccoli head. This additional pixel annotation is time consuming and expensive. To reduce annotation time and costs, it is important to have a method that can maximise the performance of the CNN with as few image annotations as possible.

Active learning is a method that can achieve this goal \citep{ren2020}. In active learning, the most informative images are automatically selected from a large pool of unlabelled images. The most informative images are then annotated manually or semi-automatically, and used for training the CNN. The hypothesis is that the generalisation performance of the CNN significantly improves when the training is done on the most informative images, because these are expected to have a higher information content for CNN optimisation \citep{ren2020}. Because only the most informative images need to be annotated with active learning, the annotation effort can be reduced while maintaining or improving the performance of the CNN.

In CNN-based active learning, pool-based sampling is the most commonly used method to select the informative images \citep{ren2020}. With pool-based sampling, each unlabelled image is first analysed, and then a set of images of a fixed size is selected to train the CNN. The image selection is carried out with a sampling method. Commonly used sampling methods are diversity sampling, uncertainty sampling, and hybrid sampling (which combines diversity and uncertainty sampling) \citep{ren2020}. With diversity sampling, images are selected that represent the diversity that exists in the set of unlabelled images. With uncertainty sampling, images are selected about which the CNN is most uncertain.

To date, most active learning methods have been developed for image classification, semantic segmentation, and object detection \citep{ren2020}. There have also been two studies on active learning for agricultural purposes. \citet{zahidi2021} researched active learning for image classification of crops and weeds, and it was shown that with active learning only 60\% of the images were needed to achieve a performance comparable to that of a CNN trained on the complete dataset. \citet{chandra2020} researched active learning for object detection in cereal crops, and it was found that 50\% of the annotation time could be saved by active learning. Unfortunately, for the broccoli harvesting robot, the use of image classification or object detection is insufficient, as for the size estimation there also needs to be a segmentation of the pixels belonging to each broccoli head. This task requires a different CNN: an instance segmentation algorithm. 

For instance segmentation algorithms, only three active learning methods have been presented \citep{lopezgomez2019, dijk2019, wang2020}. These three active learning methods were all developed for the Mask Region-based CNN (Mask R-CNN) \citep{maskrcnn}, and all methods used uncertainty sampling. The first active learning method for Mask R-CNN was developed by \citet{dijk2019}, and used the probabilistic active learning (PAL) method of \citet{krempl2014}. With PAL, the expected performance gain was calculated for the unlabelled images, and the images with the highest gain were selected for retraining. The second active learning method for Mask R-CNN was developed by \citet{wang2020}, and used a learning loss method. With this method, three additional loss prediction modules were added to the Mask R-CNN network to predict the loss of the class, box and mask of the unlabelled images. Images with a high loss were selected and labelled in a semi-supervised way using the model's output. The third active learning method for Mask R-CNN was developed by \citet{lopezgomez2019}. In this work, the image sampling was done with Monte-Carlo dropout. Monte-Carlo dropout was introduced by \citet{gal2016} and it is a frequently used sampling technique in active learning, because of its straightforward implementation. In this method, the image analysis is performed with dropout. Dropout leads to a random disconnection of some of the neurons of the CNN, forcing it to make the decision with another set of network weights. When the same image is analysed multiple times with dropout, the model output may differ between the different analyses. If the model output deviates, there seems to be uncertainty about the image, indicating that it can be a candidate for selection and annotation. 

In the work of \citet{lopezgomez2019}, two uncertainty values were calculated: the semantic uncertainty and the spatial uncertainty. The semantic uncertainty expressed the (in)consistency of Mask R-CNN to predict the class labels on an instance. The spatial uncertainty expressed the (in)consistency of Mask R-CNN to segment the object pixels of an instance. In the research of \citet{morrison2019}, an even more comprehensive uncertainty calculation was proposed for Mask R-CNN. In this work, three uncertainty values were calculated: the semantic uncertainty, the spatial uncertainty and the occurrence uncertainty. The occurrence uncertainty expressed the (in)consistency of Mask R-CNN to predict instances on the same object during the repeated image analysis. \citet{morrison2019} showed an improved predictive uncertainty of Mask R-CNN when combining the three uncertainty values into one hybrid value, compared to using the three uncertainty values separately.  Unfortunately, the uncertainty calculation of \citet{morrison2019} has not yet been applied in active learning.

The goal of our research was to develop a new active learning framework that could be used to optimise Mask R-CNN for the detection of broccoli diseases with fewer image annotations. The active learning framework was based on the uncertainty calculation of \citet{morrison2019}, but changes were made to the semantic certainty calculation to make the active learning more suitable for use on datasets with visually similar classes (like our broccoli dataset). 

We hypothesised that by using uncertainty-based active learning, the performance of Mask R-CNN can be improved faster and thereby the annotation effort can be reduced compared to a random sampling method. This hypothesis was tested on a dataset containing 16,000 images of field-grown broccoli (\textit{Brassica oleracea} var. \textit{italica}). The broccoli dataset contained images of healthy, diseased and defective broccoli heads. The first contribution of our research is a new active learning framework that can reduce annotation effort for training Mask R-CNN. The framework, which is named MaskAL, is the first active learning method that integrates three metrics to calculate the uncertainty of the instance segmentations. The MaskAL software is publicly available on \url{https://github.com/pieterblok/maskal}. The second contribution of our work is a quantitative analysis of the effects of four active learning parameters on the Mask R-CNN performance.
\newpage

\clearpage
\section{Materials and methods}
\label{materials_methods}
This section is divided into three paragraphs. Paragraph \ref{image_dataset} describes the image dataset that was used for training and evaluation. Paragraph \ref{MaskAL} describes the implementation of MaskAL. Paragraph \ref{Experiments} describes the experiments that were conducted to compare the performance between the active learning and the random sampling. 

\subsection{Image dataset}
\label{image_dataset}

\subsubsection{Broccoli images}
\label{image_acquisition}
\begin{figure}[hbt!]
  \centering
  \subfloat[] {\includegraphics[width=0.45\textwidth]{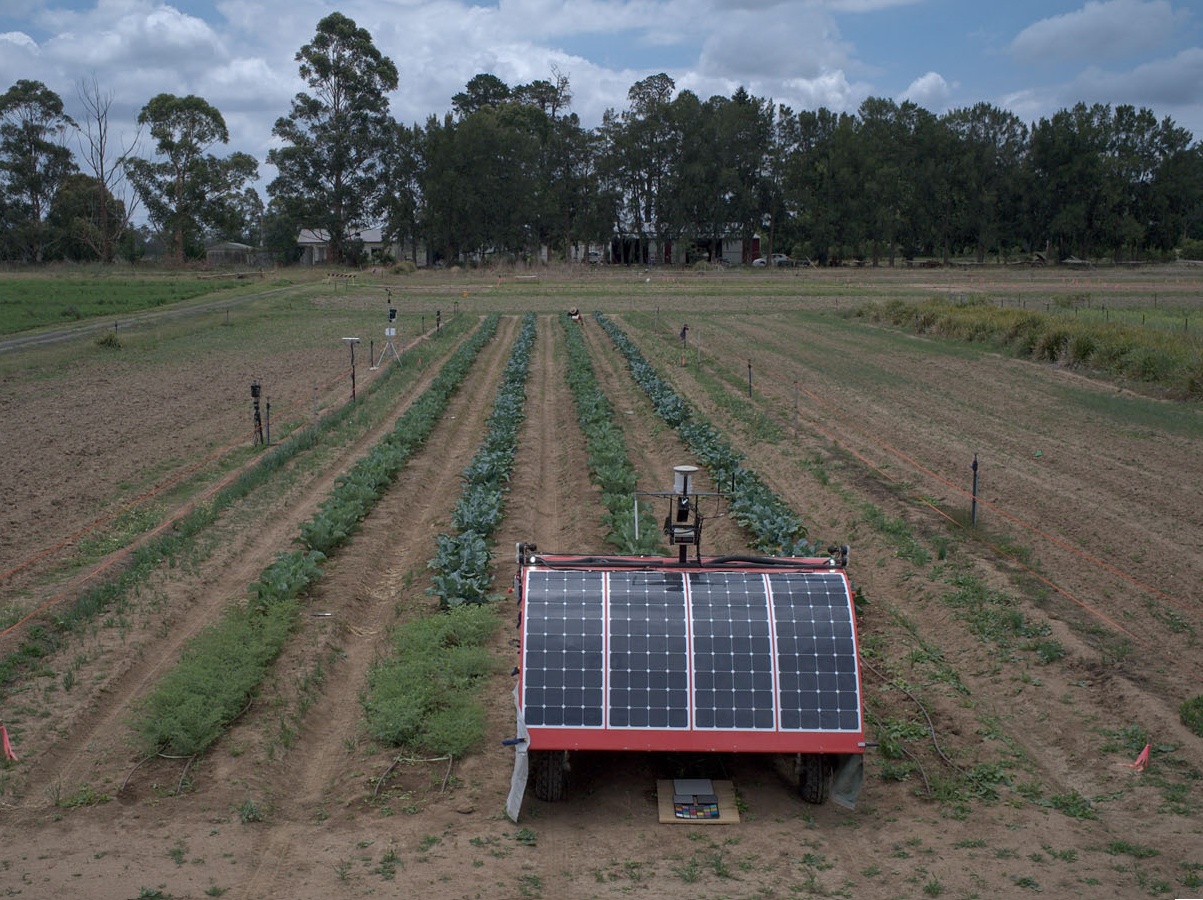}\label{fig:ladybird}}
  \hfill
  \subfloat[] {\includegraphics[width=0.45\textwidth]{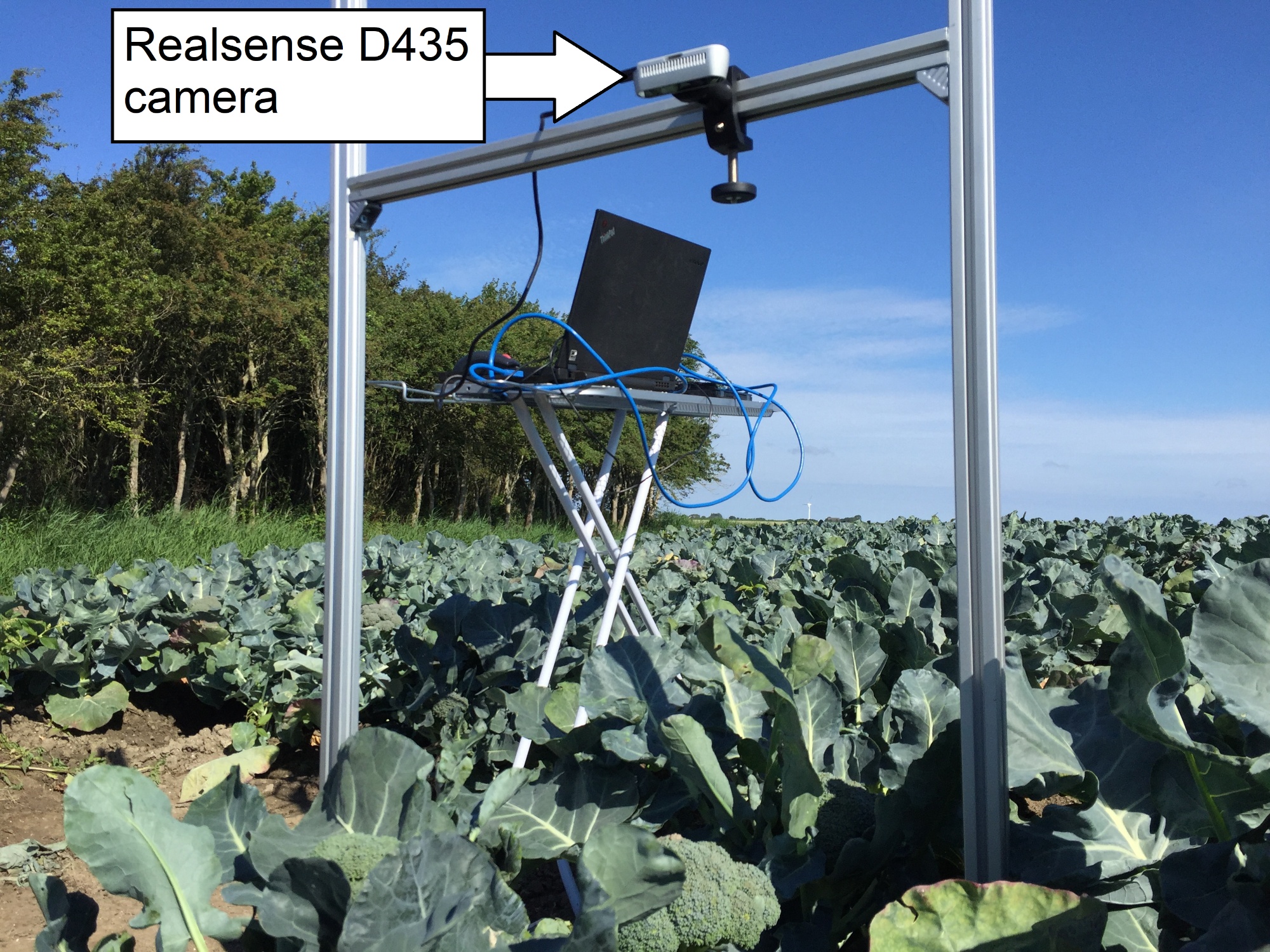}\label{fig:nl}}
  \hfill
  \subfloat[] {\includegraphics[width=0.45\textwidth]{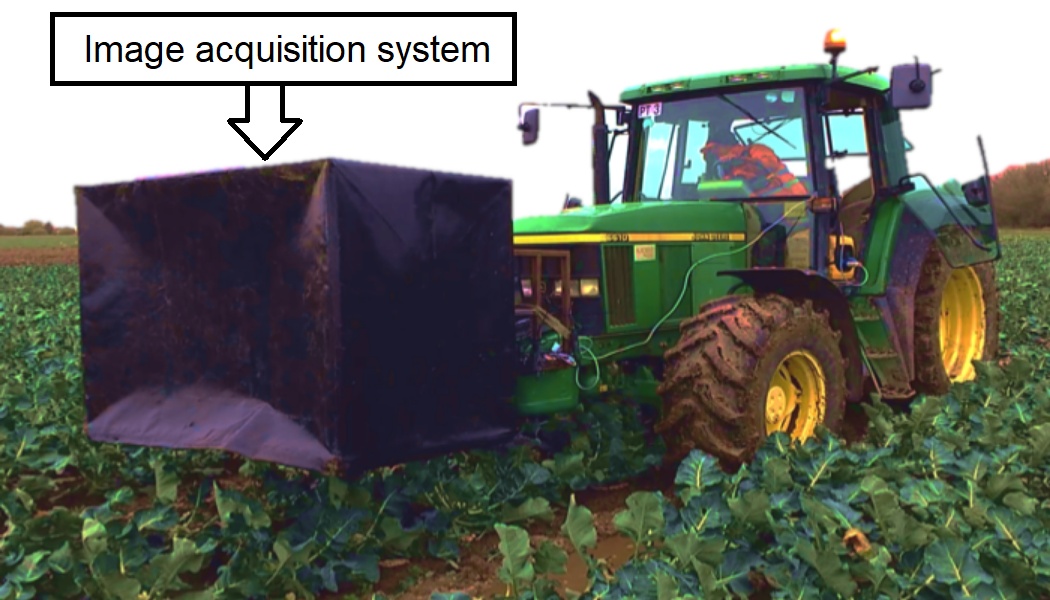}\label{fig:uk}}
  \hfill
  \subfloat[] {\includegraphics[width=0.45\textwidth]{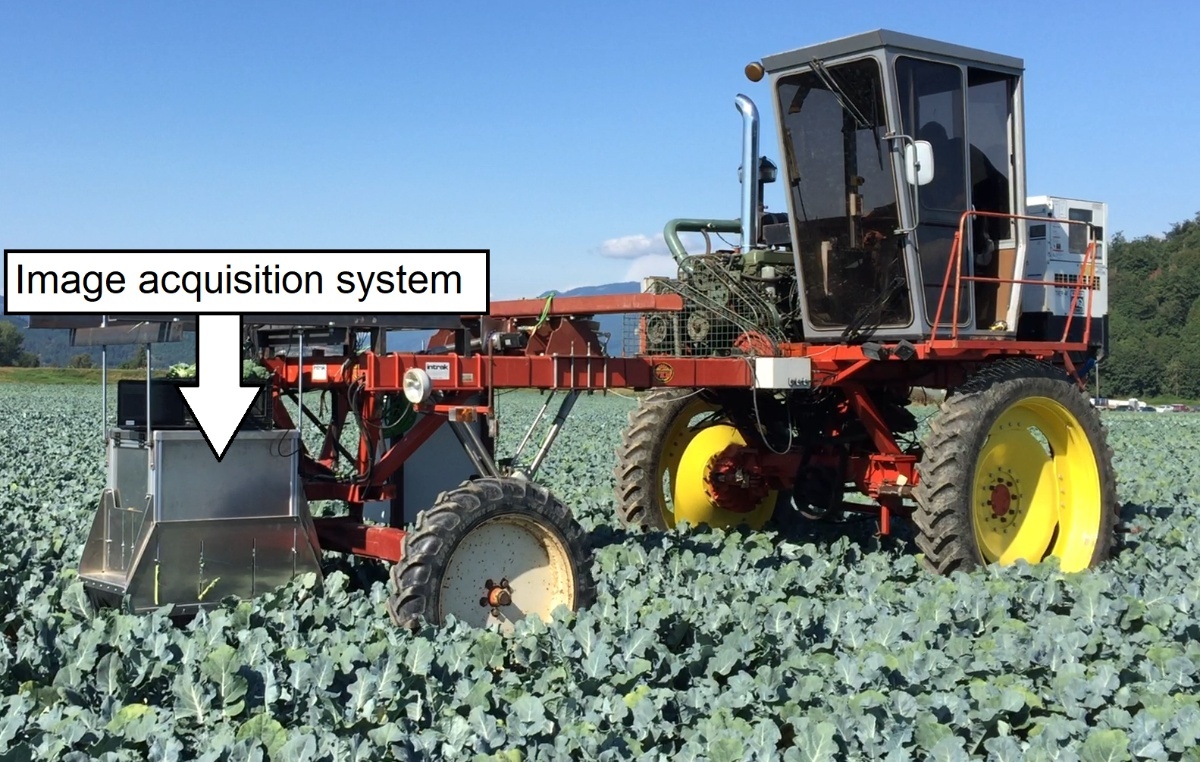}\label{fig:usa}}
  \hfill
  \caption{Overview of the image acquisition systems that were used to acquire the broccoli dataset. (a) With the Ladybird robot, broccoli images were acquired in Cobbitty (Australia) in 2017. The displayed image is from \citet{bender2020}. (b) With a stationary camera setup, broccoli images were acquired in Sexbierum (the Netherlands) in 2020. The displayed image is from \citet{blok2021_sizing}. (c) With a tractor-mounted acquisition box, broccoli images were acquired in Surfleet (the United Kingdom) in 2015. The displayed image is from \citet{kusumam2017}. (d) With a broccoli harvesting robot, broccoli images were acquired in the Netherlands and the United States of America in the period 2014-2021. The displayed image is from \citet{blok2021_sizing}.}
  \label{fig:acquisition}
\end{figure}

Our dataset consisted of 16,000 red, green, blue (RGB) colour images of field-grown broccoli. 4622 of the 16,000 images were downloaded from three online available broccoli image datasets \citep{benderdata, blok2021data, kusumamdata}. The images in these datasets were acquired with a Ladybird robot (Figure \ref{fig:ladybird}), a stationary camera frame (Figure \ref{fig:nl}) and a tractor-mounted acquisition box (Figure \ref{fig:uk}). The other 11,378 images were acquired with an image acquisition system that was attached to a broccoli harvesting robot (Figure \ref{fig:usa}) \citep{blok2021}. The detailed information about the broccoli fields, crop conditions, and camera systems can be found in Table \ref{tab:table_dataset}. 

\begin{figure}[hbt!]
  \centering
  \subfloat[] {\includegraphics[width=3.25cm, height=3.25cm]{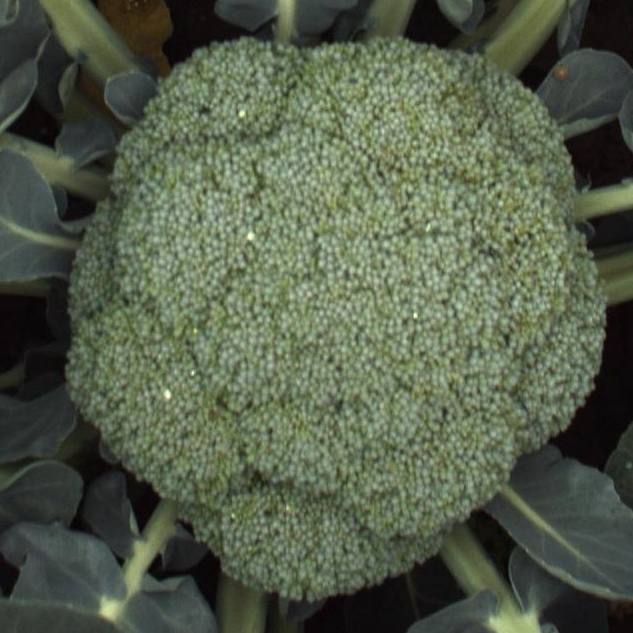}\label{fig:class_healthy}}
  \hfill
  \subfloat[] {\includegraphics[width=3.25cm, height=3.25cm]{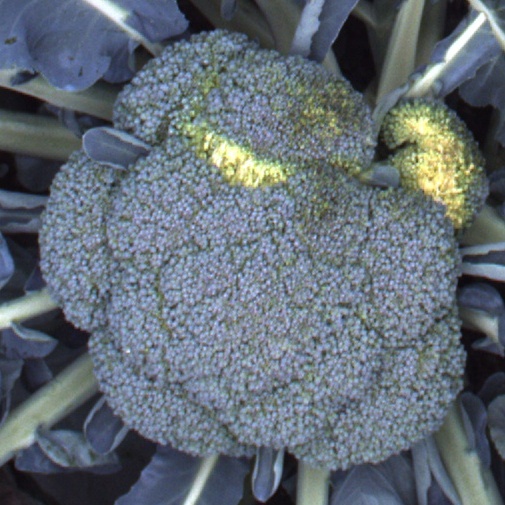}\label{fig:class_damaged}}
  \hfill
  \subfloat[] {\includegraphics[width=3.25cm, height=3.25cm]{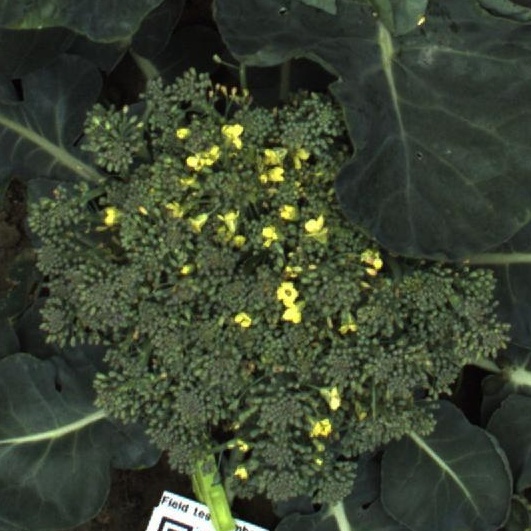}\label{fig:class_matured}}
  \hfill
  \subfloat[] {\includegraphics[width=3.25cm, height=3.25cm]{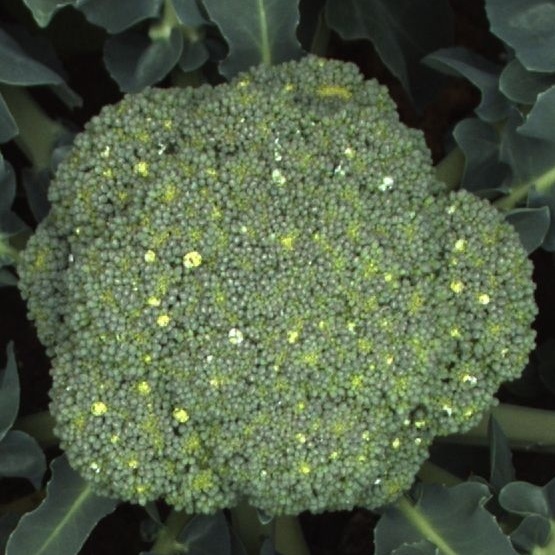}\label{fig:class_cateye}}
  \hfill
  \subfloat[] {\includegraphics[width=3.25cm, height=3.25cm]{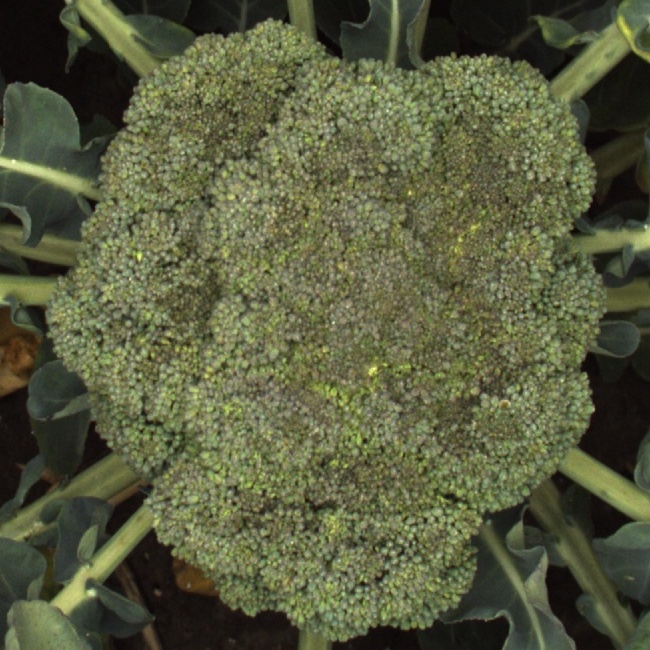}\label{fig:class_headrot}}
  \hfill
  \caption{Examples of the five broccoli classes that were present in our dataset: (a) a healthy broccoli head. (b) a damaged broccoli head. (c) a matured broccoli head. (d) a broccoli head with cat-eye.  (e) a broccoli head with head rot. The displayed images were all cropped from a bigger field image.}
  \label{fig:classes}
\end{figure}

The 16,000 images contained a total of 31,042 broccoli heads. The majority of the broccoli heads (25,704) were healthy, see one example in Figure \ref{fig:class_healthy}. The remaining 5338 broccoli heads were either diseased or defective. 1358 broccoli heads were damaged, see one example in Figure \ref{fig:class_damaged}. Broccoli heads can be damaged when they are hit by farm machinery or human harvesters. A damaged broccoli head cannot be sold for the fresh market, but it can be cut into florets for freezing, preserving some of its economic value. 1318 broccoli heads began to flower, making them unsalable. This maturation can happen when the broccoli head stays too long in the field, see one example in Figure \ref{fig:class_matured}. 1303 broccoli heads had cat-eye, which is characterised by the yellowing of some broccoli florets due to fluctuating temperatures, see one example in Figure \ref{fig:class_cateye}. Cat-eye makes the broccoli head unsaleable. 1359 broccoli heads had head rot, which is a disease that can cause necrosis or rotting of broccoli florets, making the head unsaleable, see one example in Figure \ref{fig:class_headrot}. It is important to detect head rot as soon as possible to prevent a further spread of the disease. For the application of selective harvesting and disease treatment, it was important that Mask R-CNN could distinguish between the healthy broccoli heads and those with a specific disease or defect.

\begin{table}[hbt!]
\caption{The 16,000 broccoli images were acquired on 26 fields in four countries: Australia (AUS), The Netherlands (NL), United Kingdom (UK) and United States of America (USA). The column "Acq. days" lists the number of image acquisition days performed on that particular field. The column "Ref." refers to the reference of the publicly available image dataset.}
\begin{center}
\begin{filecontents*}{csv/field_information.csv}
Field,Year,Place,ImgNum,MeasDays,Cultivar,Camera,Cultivation,Source
1,2014,Oosterbierum (NL),1105,2,Steel,1,Row,
2,2015,Oosterbierum (NL),286,3,Ironman,1,Row,
3,2015,Sexbierum (NL),480,3,Steel,1,Row,
4,2015,Surfleet (UK),2122,1,Ironman,2,Bed,A
5,2016,Sexbierum (NL),270,2,Ironman,1,Row,
6,2016,Oosterbierum (NL),206,1,Steel,1,Row,
7,2016,Sexbierum (NL),415,3,Steel,1,Row,
8,2016,Sexbierum (NL),646,2,Steel,1,Row,
9,2016,Sexbierum (NL),168,1,Steel,1,Row,
10,2017,Cobbitty (AUS),915,2,Unknown,3,Row,B 
11,2017,Sexbierum (NL),256,4,Ironman,1,Row,
12,2017,Oosterbierum (NL),481,2,Ironman,1,Row,
13,2017,Oude Bildtzijl (NL),149,1,Unknown,1,Row,
14,2018,Santa Maria (USA),2180,11,Avenger,4,Bed,
15,2018,Burlington (USA),1977,8,Emerald Crown,4,Row,
16,2018,Burlington (USA),385,3,Emerald Crown,4,Row,
17,2019,Santa Maria (USA),85,3,Eiffel,5,Bed,
18,2019,Guadalupe (USA),233,5,Avenger,4 \& 5,Bed,
19,2019,Mount Vernon (USA),660,2,Green Magic  ,4,Row,
20,2019,Mount Vernon (USA),619,1,Green Magic  ,4,Row,
21,2020,Burlington (USA),120,1,Emerald Crown,4,Row,
22,2020,Sexbierum (NL),565,1,Ironman,5,Row,C
23,2020,Sexbierum (NL),1020,1,Ironman,5,Row,C
24,2021,Blythe (USA),344,1,Unknown,6,Bed,
25,2021,Santa Maria (USA),119,1,Eiffel,6,Bed,
26,2021,Basin City (USA),194,1,Green Magic  ,6,Row,
\end{filecontents*}
\csvreader[tabular= {p{1.0cm}>{\centering}p{1.0cm}>{\centering}p{3.5cm}>{\centering}p{1.5cm}>{\centering}p{1.0cm}>{\centering}p{2.5cm}>{\centering}p{1.0cm}>{\centering\arraybackslash}p{0.7cm}},
    table head=\hline Field & Year & Place (country) & Total images & Acq. days & Broccoli cultivar & Camera & Ref. \\\hline,
    late after line=\\,
    late after last line={\\\hline}]
{csv/field_information.csv}{Field=\field, Year=\year, Place=\place, ImgNum=\imgnum, MeasDays=\days, Cultivar=\cultivar, Camera=\camera, Source=\source}
{\field & \year & \place & \imgnum & \days & \cultivar & \camera & \source}%
\end{center}
\label{tab:table_dataset}
Camera 1: AVT Prosilica GC2450, Camera 2: Microsoft Kinect 2, Camera 3: Point Gray GS3-U3-120S6C-C, Camera 4: IDS UI-5280FA-C-HQ, Camera 5: Intel Realsense D435, Camera 6: Framos D435e\\
Reference A: \citet{kusumamdata}, Reference B: \citet{benderdata}, Reference C: \citet{blok2021data}\\
\end{table}

\subsubsection{Training pool, validation set and test set}
\label{image_split}
Because the images were taken primarily on moving machines, it could happen that a unique broccoli head was photographed several times. To avoid the training, validation, or test set containing images of the same broccoli head, we first grouped the image frames of unique broccoli heads. This grouping was done after converting the global navigation satellite system (GNSS) coordinates of the tractor to the location of each broccoli head in the image so that they could be identified and separated. The image frames belonging to a unique broccoli head were placed into either the training pool, validation set, or test set. 

The training pool consisted of 14,000 images, and these images could be used to train Mask R-CNN and to sample new images. In the training pool, there were 27,009 broccoli heads of which most were healthy (the class distribution was 27:1 (healthy : disease/defect)). The validation set consisted of 500 images and 1020 broccoli heads. The validation images were used during the training process to check whether Mask R-CNN was overfitting. The validation images were selected with an algorithm that prioritised the selection of images with diseased or defective broccoli heads over images with only healthy broccoli heads. As a result, the class distribution in the validation set was 5:1 (healthy : disease/detect). With this class distribution, we were better able to evaluate the Mask R-CNN performance on the five broccoli classes.

Instead of one test set, we used three test sets of 500 images each. The three test sets contained respectively 961, 1009, and 1043 broccoli heads. The three test sets were completely independent of the training process, and each test set served as an independent image set for each of our three experiments (refer to paragraph \ref{Experiments}). Because the outcome of an experiment influenced the parameter choice in the next experiment, new test sets were needed that were independent of the previously used test set. The image selection of the three test sets was performed with the same algorithm that prioritised the selection of images with diseased or defective broccoli heads. The resulting class distribution in the three test sets was 5:1 (healthy : disease/detect). 

\subsection{MaskAL}
\label{MaskAL}
The MaskAL procedure consisted of four sequential steps. First, a subset of images was selected from the training pool, and annotated. Second, Mask R-CNN was trained on these images. Third, the trained Mask R-CNN model was evaluated on the independent test set to determine its performance. Fourth, a new subset of images was selected from the training pool with either random sampling or uncertainty sampling (the active learning). After the fourth step, the selected images were annotated and added to the previous training set. Mask R-CNN was retrained on this combined set of images, after which it was evaluated and used to sample new images. This entire procedure was repeated for a number of sampling iterations. The MaskAL procedure is explained in more detail in the pseudo-code of Algorithm \ref{algo1} (the blue-coloured functions highlight the four consecutive steps).

\begin{figure}[hbt!]
  \centering
    \includegraphics[width=1\textwidth]{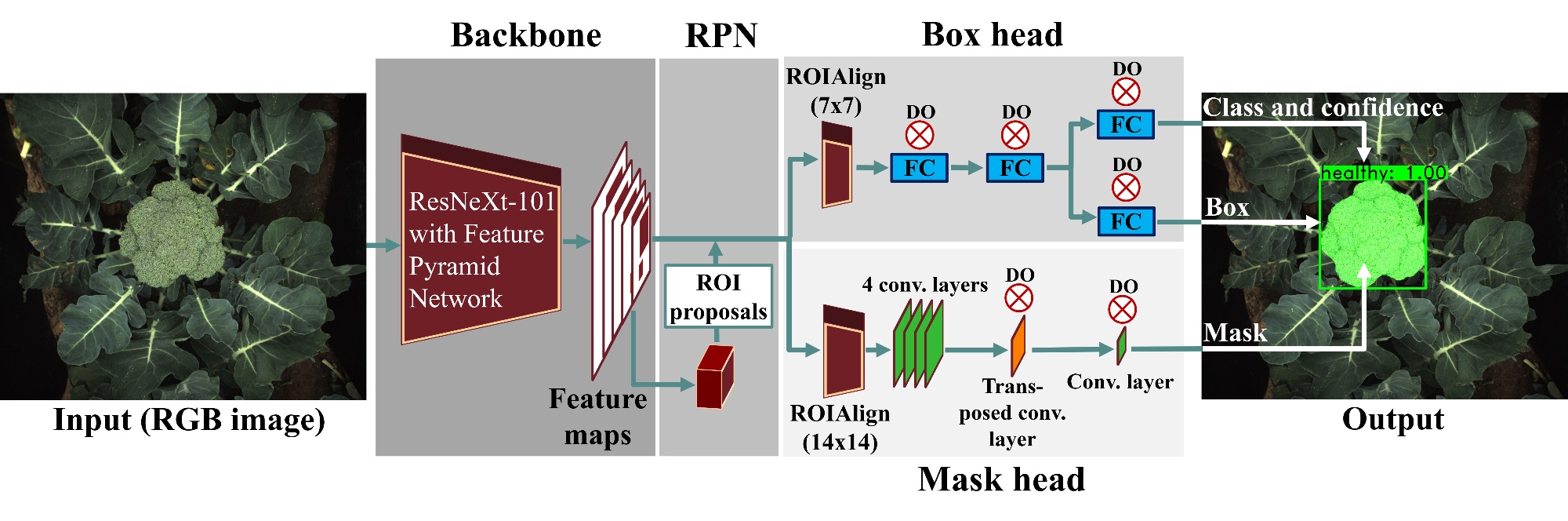}
    \caption{Schematic representation of the Mask R-CNN network architecture in MaskAL. The white circles with the red crosses indicate the network layers with dropout. Conv., DO, RPN, ROI and FC, are abbreviations of respectively convolutional, dropout, region proposal network, region of interest, and fully connected layers. The numbers between brackets give the output dimensions of the ROIAlign layer. The image was adapted from \citet*{shi2019}.}
    \label{fig:MaskAL_architecture}
\end{figure}

MaskAL was built as a software shell on top of the Mask R-CNN code of Detectron2 (version 0.4) \citep{detectron2}. In our research, Mask R-CNN was equipped with the ResNeXt-101 (32x8d) backbone \citep{resnext2017}. Before the training and sampling could be performed, dropout had to be applied to several network layers of Mask R-CNN. In the box head of Mask R-CNN, dropout was applied to each fully connected layer, see Figure \ref{fig:MaskAL_architecture}. This dropout placement was in line with \citet{gal2016} and would allow us to capture the variation in the predicted classes and the bounding box locations. Dropout was also applied to the last two convolutional layers in the mask head of Mask R-CNN to be able to capture the variation in the pixel segmentation, see Figure \ref{fig:MaskAL_architecture}. The severity of the dropout was made configurable in MaskAL by means of the dropout probability. The dropout probability determined the chance of neurons getting disconnected in the network layers and the value could be configured between 0.0 (no dropout) and 1.0 (complete dropout). 

% \Cref{algo1}
% \lipsum[1-5]
\begin{algorithm}
\caption{MaskAL \\The blue-coloured words are the core functions.}
\label{algo1}
\SetAlgoLined
\DontPrintSemicolon
\LinesNumbered

\SetKwData{Strategies}{strategies}
\SetKwData{SampIter}{sampling-iterations}
\SetKwData{InitDSSize}{initial-dataset-size}
\SetKwData{Poolsize}{sample-size}
\SetKwData{Trainpool}{training-pool}
\SetKwData{Valset}{val-set}
\SetKwData{Testset}{test-set}
\SetKwData{Strat}{sampling-method}
\SetKwData{StratLen}{length(strategies) $-$ 1}
\SetKwData{Strats}{strategies[\textit{s}]}
\SetKwData{InitDS}{initial-dataset}
\SetKwData{InitModel}{initial-model}
\SetKwData{TrainModel}{model}
\SetKwData{mAP}{mAP}
\SetKwData{mAPs}{mAPs}
\SetKwData{AvImg}{available-images}
\SetKwData{TrainImg}{train-images}
\SetKwData{True}{True}
\SetKwData{False}{False}
\SetKwData{MW}{prev-weights}
\SetKwData{CW}{coco-weights}
\SetKwData{Pool}{pool}
\SetKwData{Uncert}{'uncertainty'}
\SetKwData{Cert}{'certainty'}
\SetKwData{Rand}{'random'}
\SetKwData{FP}{forward-passes}
\SetKwData{IOUT}{iou-threshold}
\SetKwData{nc}{num-classes}
\SetKwData{Method}{certainty-method}
\SetKwData{AnnotTrain}{annotated-train-set}
\SetKwData{AnnotVal}{annotated-val-set}
\SetKwData{AnnotTest}{annotated-test-set}
\SetKwData{AnnotPool}{annotated-pool}
\SetKwData{NMS}{nms-threshold}
\SetKwData{CONF}{conf-threshold}
\SetKwData{DPROB}{dropout-probability}
\SetKwData{EmptyArray}{[ ]}
\SetKwData{mAPInsert}{mAPs.insert(mAP)}

\SetKwFunction{InitDSFunc}{SampleInitialDatasetRandomly}
\SetKwFunction{TrainModelFunc}{\hl{TrainMaskRCNN}}
\SetKwFunction{EvalFunc}{\hl{EvaluateMaskRCNN}}
\SetKwFunction{Uncertainty}{\hl{UncertaintySampling}}
\SetKwFunction{Certainty}{CertaintySampling}
\SetKwFunction{Random}{RandomSampling}
\SetKwFunction{Classes}{GetNumberOfClasses}
\SetKwFunction{Annotate}{\hl{Annotate}}

\SetKwInOut{Input}{Inputs}
\SetKwInOut{Output}{Outputs}
\Input{sampling-method : 'uncertainty' or 'random' \newline sampling-iterations : integer \newline sample-size : integer \newline initial-dataset-size : integer \newline dropout-probability: float \newline forward-passes : integer \newline certainty-method : 'average' or ' minimum'  \newline training-pool : dataset with all available images for training and sampling \newline val-set : validation dataset \newline test-set : test dataset}
\Output{mAPs : an array with mean average precision values of size sampling-iterations+1 \newline}

\SetKwFunction{FMain}{MaskAL}
\SetKwProg{Fn}{Function}{:}{}
\Fn{\FMain{\Strat, \SampIter, \Poolsize, \InitDSSize, \DPROB, \FP, \Method, \Trainpool, \Valset, \Testset}}
{
    \mAPs$\gets$ \EmptyArray\;
    \InitDS$\gets$ \InitDSFunc{\Trainpool, \InitDSSize}\;
    \AnnotTrain, \AnnotVal, \AnnotTest $\gets$ \Annotate{\InitDS, \Valset, \Testset}\;
    \TrainModel$\gets$ \TrainModelFunc{\AnnotTrain, \AnnotVal, \CW, \DPROB}\;
    \mAP$\gets$ \EvalFunc{\TrainModel, \AnnotTest}\;
    \mAPInsert\;
    \AvImg$\gets$ \textit{\Trainpool} $-$ \InitDS\;
    \For{$i \gets 1$ \KwTo \SampIter}{
        \uIf{\Strat = \Uncert}
            {\Pool$\gets$ \Uncertainty{\TrainModel, \AvImg, \Poolsize, \DPROB, \FP, \Method}\;}
        \uElseIf{\Strat = \Rand}
            {\Pool$\gets$ \Random{\AvImg, \Poolsize}\;}
        \AnnotPool$\gets$ \Annotate{\Pool}\;
        \AnnotTrain$\gets$ \AnnotTrain $+$ \AnnotPool\;
        \MW$\gets$ \TrainModel\;
        \TrainModel$\gets$ \TrainModelFunc{\AnnotTrain, \AnnotVal, \MW, \DPROB}\;
        \mAP$\gets$ \EvalFunc{\TrainModel, \AnnotTest}\;
        \mAPInsert\;
        \AvImg$\gets$ \AvImg $-$ \AnnotPool\;
    }
    % }
    \KwRet \mAPs\;
}
\end{algorithm}

\subsubsection{Step 1 - Annotate}
\label{annotate} 
The first step in MaskAL involved the annotation of the selected images. At algorithm initialisation, this annotation was done on a subset of images that was randomly sampled from the training pool. The images were annotated by two crop experts, who used the LabelMe software (version 4.5.6) \citep{labelme2016}. In our research, all images were annotated beforehand, as this allowed us to conduct three experiments without being interrupted for doing the image annotations. In addition, by annotating all images, we were able to train Mask R-CNN on the entire training pool (see paragraph \ref{experiment3}).

\subsubsection{Step 2 - Train Mask R-CNN}
\label{Training}
After the image annotation, Mask R-CNN was trained on the selected images. The training procedure was identical for the uncertainty sampling and the random sampling. 

The training was performed with a learning rate of 1.0$\cdot$10\textsuperscript{-2}, and an image batch size of two. The stochastic gradient descent optimiser was used with a momentum of 0.9 and a weight decay of 1.0$\cdot$10\textsuperscript{-4}. During training, two data augmentations were employed. The first augmentation was a random horizontal flip of the image with a probability of 0.5. The second augmentation was an image resizing along the shortest edge of the image while maintaining the aspect ratio of the image. The total number of training iterations was proportional to the number of training images: for each multiple of 500 training images, 2,500 training iterations were added to the base number of 2,500 iterations. The training was performed with dropout as a regularisation technique to enhance the generalisation performance. This procedure was in line with the training procedure of \citet{gal2016}.

At MaskAL initialisation, the network weights of Mask R-CNN were initialised with the weights of a Mask R-CNN model that was pretrained on the Microsoft Common Objects in Context (COCO) dataset \citep{lin2014}. After the first training procedure, the transfer learning was done with the weights of the previously trained Mask R-CNN model to minimise the effects of catastrophic forgetting.

In our training pool, there was a severe class imbalance (healthy : disease/defect = 27:1). Due to this class imbalance, the random sampling was more likely to sample images with healthy broccoli heads than images with damaged, matured, cat-eye or head rot broccoli heads. This could eventually lead to a much worse performance than the uncertainty sampling. To prevent that our comparison would be too much influenced by the class imbalance, it was decided to train Mask R-CNN with a data oversampling strategy \citep{gupta2019lvis}. With this strategy, a specific image was repeatedly trained by Mask R-CNN if that image contained a minority class (the minority classes were damaged, matured, cat-eye and head rot). By repeating the images with minority classes during the training, this oversampling strategy was expected to reduce the negative effect of the class imbalance on the Mask R-CNN performance.

\subsubsection{Step 3 - Evaluate Mask R-CNN}
\label{evaluation} 
After the training, the Mask R-CNN model was evaluated on the independent test set to determine its performance. The performance metric was the mean average precision (mAP), which expressed the classification and instance segmentation performance of Mask R-CNN. A mAP value close to zero indicated an incorrect classification and/or inaccurate instance segmentation, while a value close to 100 indicated a correct classification and accurate instance segmentation. 

The evaluation of Mask R-CNN was done without dropout, and with a fixed threshold of 0.01 on the non-maximum suppression (NMS). This NMS threshold removed all instances that overlapped at least 1\% with a more confident instance, essentially meaning that only one instance segmentation was done on an object. This approach was considered valid since the broccoli heads grew solitary and did not overlap each other in the images. 

\subsubsection{Step 4 - Uncertainty sampling}
\label{uncertainty_sampling}
After the training and evaluation, new images were sampled from the training pool. With the random sampling method, this was done randomly. With the uncertainty sampling, images were sampled about which the trained Mask R-CNN model was most uncertain. The uncertainty sampling involved four steps: the image analysis with Monte-Carlo dropout (paragraph \ref{procedure}), the grouping of the instance segmentations into instance sets (paragraph \ref{instance_sets}), the calculation of the certainty values (paragraph \ref{certainty_calculation}), and the image sampling (paragraph \ref{sampling}).

\begin{figure}[hbt!]
  \centering
  \subfloat[] {\includegraphics[width=0.495\textwidth]{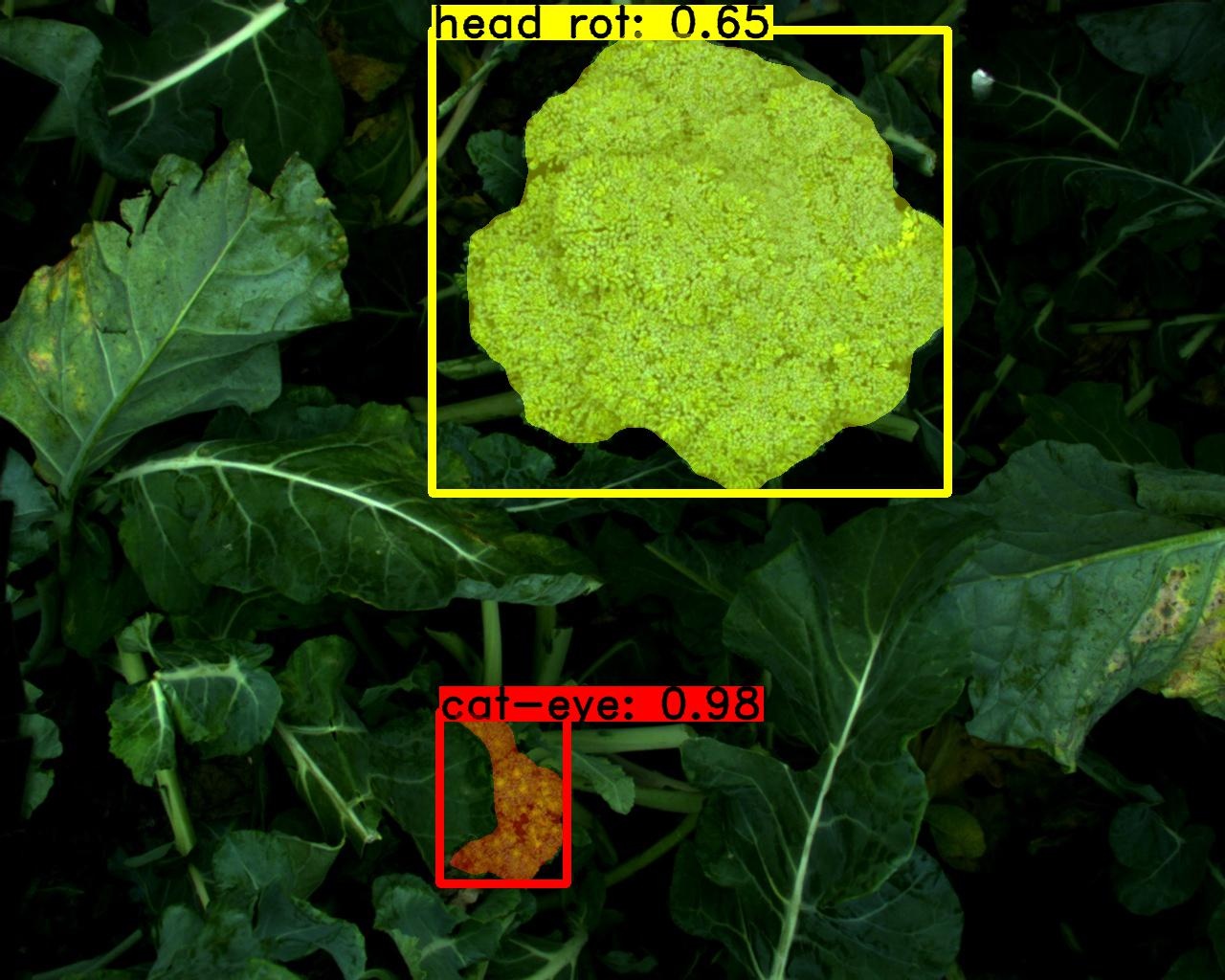}\label{fig:mc1}}
  \hfill
  \subfloat[] {\includegraphics[width=0.495\textwidth]{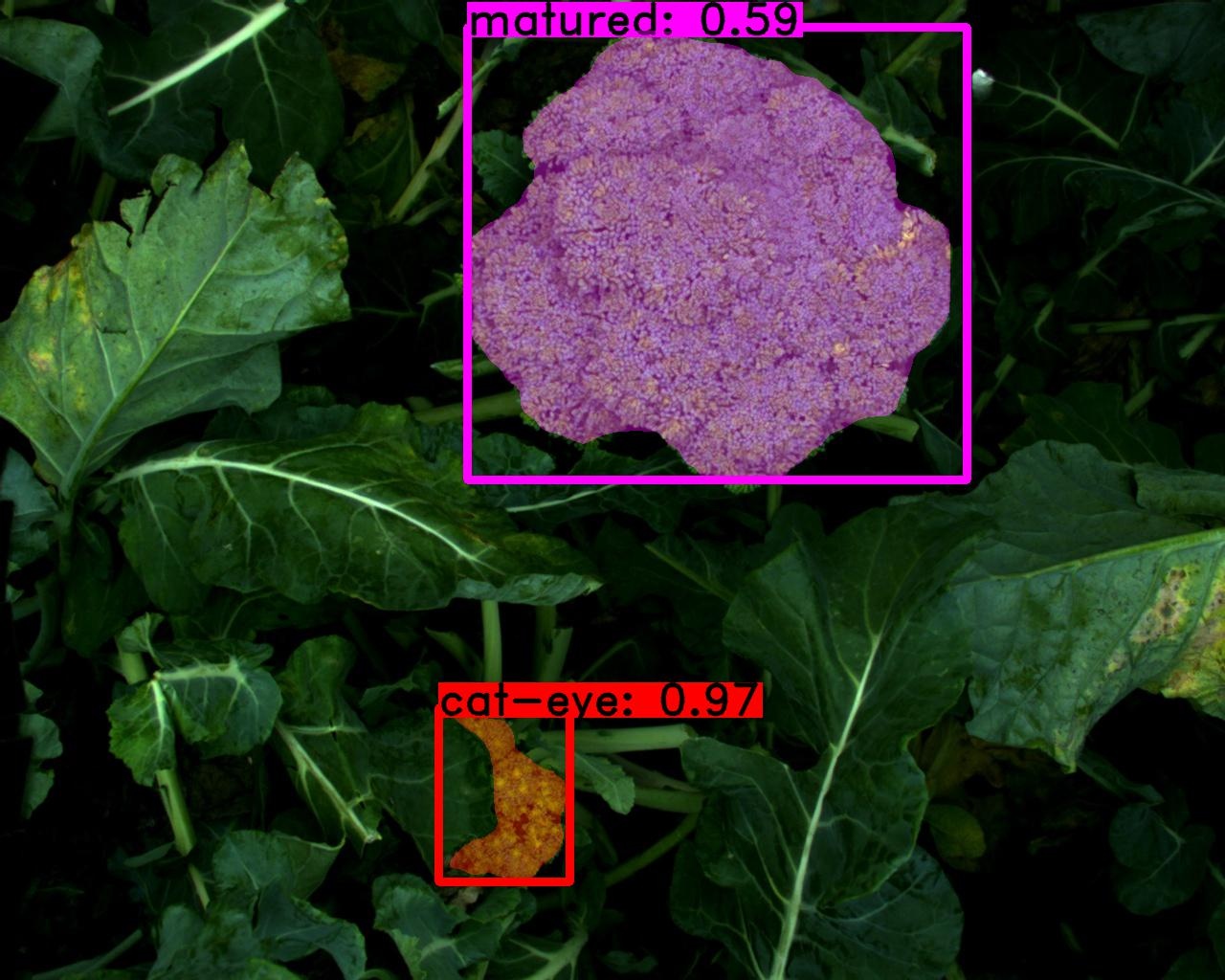}\label{fig:mc2}}
  \hfill
  \subfloat[] {\includegraphics[width=0.495\textwidth]{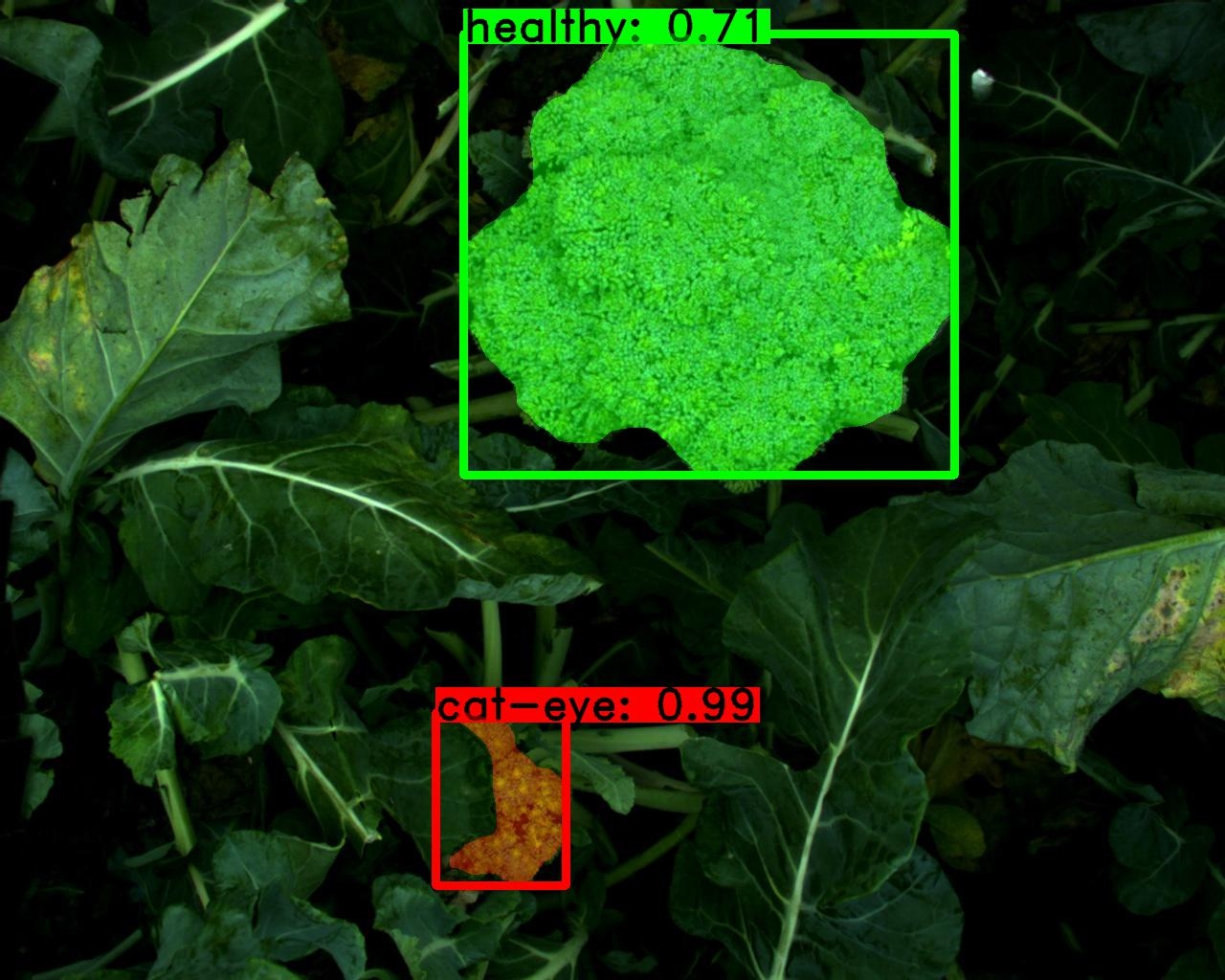}\label{fig:mc3}}
  \hfill
  \subfloat[] {\includegraphics[width=0.495\textwidth]{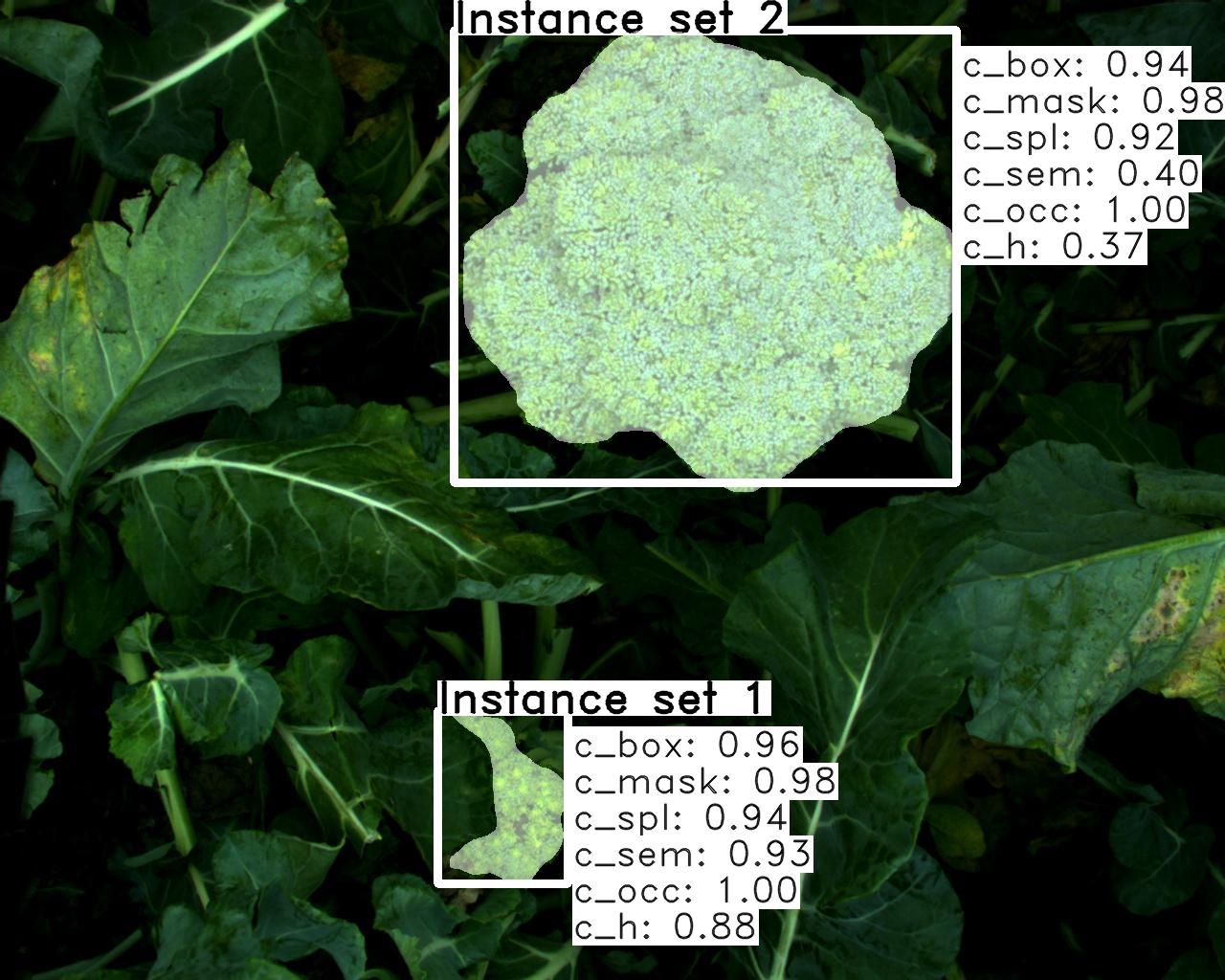}\label{fig:obs}}
  \hfill
  \caption{A visual example of the Monte-Carlo dropout method and the calculation of the certainty values. (a) After the first forward pass with dropout, Mask R-CNN produced two instances: one instance of class cat-eye with a high confidence score (0.98) and one instance of class head rot with a lower confidence score (0.65). (b) The same image was analysed again during a second forward pass, resulting a confident cat-eye instance (0.97) and a less confident matured instance (0.59). (c) After the third forward pass, Mask R-CNN produced a confident cat-eye instance (0.99) and a moderately confident healthy instance (0.71). (d) After the three forward passes, the instance segmentations were grouped into two instance sets, based on spatial similarity. An instance set is a group of different instance segmentations that appear on the same broccoli head. The white bounding box and mask of the instance sets represent the average box and mask of the instance segmentations. On each instance set, three certainty values were calculated: the semantic certainty ($c_\texttt{sem}$), the occurrence certainty ($c_\texttt{occ}$) and the spatial certainty ($c_\texttt{spl}$), which was the product of the spatial certainty of the bounding box ($c_\texttt{box}$) and the mask ($c_\texttt{mask}$). The certainty of the instance set ($c_\texttt{h}$) was calculated by multiplying the $c_\texttt{sem}$, $c_\texttt{occ}$ and $c_\texttt{spl}$.}
  \label{fig:uncertainty}
\end{figure}

\paragraph{Step 4.1 - Monte-Carlo dropout}
\label{procedure}
Each available image from the training pool was analysed with the Monte-Carlo dropout method. A user-specified number of forward passes determined how many times the same image was analysed with the trained Mask R-CNN model. During the repeated image analysis, the dropout caused the random disconnection of some of the neurons in the head branches of Mask R-CNN. This random neuron disconnection could lead to different model outputs, see Figure \ref{fig:mc1}, \ref{fig:mc2} and \ref{fig:mc3}. The repeated image analysis was done with a fixed threshold of 0.01 on the non-maximum suppression and a fixed threshold of 0.5 on the confidence level. 

\paragraph{Step 4.2 - Instance sets}
\label{instance_sets}
After the repeated image analysis, the outputs of Mask R-CNN were grouped into instance sets. An instance set is a group of instance segmentations from multiple forward passes that appear on the same object in the image, see two examples in Figure \ref{fig:obs}. On these instance sets, the certainty values were calculated. 

The grouping of the instance segmentations into instance sets was based on the spatial similarity method of \citet{morrison2019}. The method added an instance segmentation to an instance set when the intersection over union (IoU) between this segmentation and at least one other segmentation from the instance set exceeded a certain threshold, $\tau_\texttt{IoU}$. The IoU is a metric for the spatial overlap between two mask segmentations, $M_1$ and $M_2$, and the value varies between zero (no overlap) and one (complete overlap), see Equation \ref{eq_iou}. 

\begin{equation}
\label{eq_iou}
\texttt{IoU}(M_1, M_2) = \frac{|M_1 \cap M_2|}{|M_1 \cup M_2|} \quad \quad \mbox{where $|\cdot|$ gives the total number of mask pixels} 
\end{equation}

A new instance set was created when the segmentation did not exceed the $\tau_\texttt{IoU}$ threshold. It was assumed that this segmentation then represented a different object. In our research, the $\tau_\texttt{IoU}$ was set to 0.5 and this value was adopted from \citet{morrison2019}. 

\paragraph{Step 4.3 - Certainty calculation}
\label{certainty_calculation}
Three certainty values were calculated for each instance set: the semantic certainty, the spatial certainty, and the occurrence certainty. The certainty calculations were adopted from \citet{morrison2019}, but changes were made to the semantic certainty calculation to make it more suitable for use on datasets with visually similar classes (like our broccoli dataset). 

The semantic certainty, $c_\texttt{sem}$, was a measure of the consistency of Mask R-CNN to predict the class labels within an instance set. The $c_\texttt{sem}$ value was close to zero when there was a low semantic certainty, and close to one when there was a high semantic certainty. \citet{morrison2019} calculated the $c_\texttt{sem}$ value by taking the difference between the average confidence score of the first and the second most probable class of the instance set. This method is known as margin sampling, but a disadvantage is that it does not take into account the confidence scores of the less probable classes. This is undesirable when the dataset has a high degree of inter-class similarity, because then Mask R-CNN has a tendency to hesitate between more than two classes. This multi-class hesitation was also expected in our broccoli dataset. To overcome the disadvantage of the margin sampling, the $c_\texttt{sem}$ calculation was upgraded with an entropy-based equation, which took the confidence scores, $P$, of all classes, \textit{K} = $\set{k_1,\ldots,k_n}$, into account, see Equation \ref{eq_entropy}. However, with the entropy calculation, the $c_\texttt{sem}$ value would be low when there was class certainty, and this was opposite of Morrison's $c_\texttt{sem}$ value. Also, the entropy calculation could result in values higher than one, which deviated from Morrison's $c_\texttt{sem}$ value that was bound between zero and one. With two additional calculations these issues were solved. First, the entropy value, $H$, was divided by the maximum entropy value, $H_\texttt{max}$, so that the resulting value was bound between zero and one (see Equation \ref{eq_usem_instance}). The $H_\texttt{max}$ value was calculated with Equation \ref{eq_entropy_max}, and this value represented a situation where the confidence scores of all classes were equal, which was the case when Mask R-CNN had the lowest certainty in predicting the class labels. Then, the resulting value from the division was inverted, such that a high $H_\texttt{sem}$ value would result when there was a high semantic certainty. With Equation \ref{eq_usem}, the $H_\texttt{sem}$ values of all instances belonging to an instance set, \textit{S} = $\set{s_1,\ldots,s_r}$, were averaged. This resulted in one semantic certainty value, $c_\texttt{sem}$, per instance set. Figure \ref{fig:obs} visualises two estimations of $c_\texttt{sem}$. 

\begin{equation}
\label{eq_entropy}
\begin{split}
\texttt{H}(K) & = -\sum\limits_{i=1}^{n} P(k_i) \cdot \log P(k_i) \quad \quad \mbox{with \textit{K} = $\set{k_1,\ldots,k_n}$}
\end{split}
\end{equation}

\begin{equation}
\label{eq_usem_instance}
\texttt{H}_\texttt{sem}(s) = 1 - \left( \frac{\texttt{H}(K)}{\texttt{H}_\texttt{max}(K)} \right) \quad \quad \mbox{where $s$ is an instance of instance set $S$} 
\end{equation}

\begin{equation}
\label{eq_entropy_max}
\begin{split}
\texttt{H}_\texttt{max}(K) & = -n \cdot \left(\tfrac{1}{n} \cdot \log \tfrac{1}{n}\right) \quad \quad \mbox{where $n$ is the number of classes} 
\end{split}
\end{equation}

\begin{equation}
\label{eq_usem}
\texttt{c}_\texttt{sem}(S) =  \frac{1}{r}  \cdot \sum\limits_{i=1}^{r}\texttt{H}_\texttt{sem}(s_i) \quad \quad \mbox{with \textit{S} = $\set{s_1,\ldots,s_r}$}
\end{equation}

The spatial certainty, $c_\texttt{spl}$, was a measure of the consistency of Mask R-CNN to determine the bounding box locations and to segment the object pixels within an instance set. The $c_\texttt{spl}$ value was close to zero when there was little spatial consistency between the boxes and the masks in the instance set, and the value was close to one when there was much spatial consistency. The $c_\texttt{spl}$ value was calculated by multiplying the spatial certainty value of the bounding box ($c_\texttt{box}$) by the spatial certainty value of the mask ($c_\texttt{mask}$), see Equation \ref{eq_uspl}. The $c_\texttt{box}$ and the $c_\texttt{mask}$ values were the mean IoU values between the average box and mask of the instance set (respectively denoted as $\bar{B}$ and $\bar{M}$) and each individual box and mask prediction within that instance set (respectively denoted as $B$ and $M$), refer to Equation \ref{eq_ubox} and \ref{eq_umask}. The average box, $\bar{B}$, was formed from the centroids of the corner points of the individual boxes in the instance set (see the white boxes in Figure \ref{fig:obs}). The average mask, $\bar{M}$, represented the segmented pixels that appeared in at least 25\% of the individual masks in the instance set (see the white masks in Figure \ref{fig:obs}). The value of 25\% was found to produce the most consistent average masks for our broccoli dataset. 

\begin{equation}
\label{eq_uspl}
\texttt{c}_\texttt{spl}(S) = \texttt{c}_\texttt{box}(S) \cdot  \texttt{c}_\texttt{mask}(S) \quad \quad \quad \mbox{with \textit{S} = $\set{s_1,\ldots,s_r}$}\\
\end{equation}

\begin{equation}
\label{eq_ubox}
\texttt{c}_\texttt{box}(S) = \frac{1}{r} \cdot \sum\limits_{i=1}^{r}\texttt{IoU}(\bar{B}(S), B(s_i)) \quad \quad \mbox{with \textit{S} = $\set{s_1,\ldots,s_r}$}\\
\end{equation}

\begin{equation}
\label{eq_umask}
\texttt{c}_\texttt{mask}(S) = \frac{1}{r} \cdot \sum\limits_{i=1}^{r}\texttt{IoU}(\bar{M}(S), M(s_i)) \quad \quad \mbox{with \textit{S} = $\set{s_1,\ldots,s_r}$}\\
\end{equation}

The occurrence certainty, $c_\texttt{occ}$, was a measure of the consistency of Mask R-CNN to predict instances on the same object during the repeated image analysis. The $c_\texttt{occ}$ value was close to zero, when there was little consensus in predicting an instance on the same object in each forward pass. The $c_\texttt{occ}$ value was one when Mask R-CNN predicted an instance on the same object in each forward pass. The $c_\texttt{occ}$ value was calculated by dividing the number of instances belonging to an instance set, $r$, by the number of forward passes, $fp$, see Equation \ref{eq_un}. 

\begin{equation}
\label{eq_un}
\texttt{c}_\texttt{occ}(S) = \frac{r}{fp} \quad \quad \quad \mbox{with \textit{S} = $\set{s_1,\ldots,s_r}$}\\
\end{equation}

The semantic, spatial, and occurrence certainty values were multiplied into one certainty value for each instance set, see Equation \ref{eq_uh}. With this multiplication, the three certainty values were considered equally important in determining the overall certainty, $\texttt{c}_\texttt{h}$, of an instance set.   

\begin{equation}
\label{eq_uh}
\texttt{c}_\texttt{h}(S) = \texttt{c}_\texttt{sem}(S) \cdot \texttt{c}_\texttt{spl}(S) \cdot \texttt{c}_\texttt{occ}(S) \quad \quad \mbox{with \textit{S} = $\set{s_1,\ldots,s_r}$}\\
\end{equation}

Because the Mask R-CNN training and testing was done on images and not on individual instances, it was needed to combine the certainties of the instance sets into one certainty value for the entire image. The image certainty value was calculated with either the average method or the minimum method. With the average method, the image certainty value was the average certainty value of all instance sets in the image, \textit{I} = $\set{S_1,\ldots,S_t}$, see Equation \ref{eq_uavg}. In Figure \ref{fig:obs}, the average certainty value was 0.62 ((0.88 + 0.37)/2). With the minimum method, the image certainty value was the lowest certainty value of all instance sets, see Equation \ref{eq_umin}. In Figure \ref{fig:obs}, the minimum certainty value was 0.37. The certainty calculation method was made configurable in MaskAL, so that we could do an experiment to assess its effect on the active learning performance (this is explained in paragraph \ref{experiment1}).  

\begin{equation}
\label{eq_uavg}
\texttt{c}_\texttt{avg}(I) = \frac{1}{t} \cdot \sum\limits_{i=1}^{t}\texttt{c}_\texttt{h}(S_i) \quad \quad \mbox{with \textit{I} = $\set{S_1,\ldots,S_t}$}
\end{equation}

\begin{equation}
\label{eq_umin}
\texttt{c}_\texttt{min}(I) = \min(\texttt{c}_\texttt{h}(I)) \quad \quad \mbox{with \textit{I} = $\set{S_1,\ldots,S_t}$}
\end{equation}

\paragraph{Step 4.4 - Sampling}
\label{sampling}
After calculating the certainty value of each image from the training pool, a subset of images was selected about which Mask R-CNN was most uncertain. The size of the image set, hereinafter referred to as the sample size, was made configurable in MaskAL. This allowed us to do an experiment to assess the effect of the sample size on the active learning performance (this is explained in paragraph \ref{experiment2}). 

\subsection{Experiments}
\label{Experiments}
Three experiments were set up, with the final objective of comparing the performance of the active learning with the performance of the random sampling. Before this comparison could be done, experiments 1 and 2 were performed to investigate how to optimise the active learning. 

\subsubsection{Experiment 1}
\label{experiment1}
The objective of experiment 1 was to test the effect of the dropout probability, certainty calculation method, and number of forward passes on the active learning performance. This test would reveal the optimal settings for these parameters, which were assumed to have the most influence on the active learning performance, because they all influenced the calculation of the certainty value of the image.

The experiment was done with three dropout probabilities: 0.25, 0.50, and 0.75. Dropout probability 0.50 was a frequently used probability in analogous active learning research, for example \citet{aghdam2019}, \citet{gal2016} and \citet{lopezgomez2019}. With this dropout probability, there was a moderate chance of dropout during the image analysis. The dropout probabilities 0.25 and 0.75 were chosen to have a lower chance of dropout and a higher chance of dropout, compared to the dropout probability 0.50. 

Two certainty calculation methods were tested: the average method and the minimum method, which are both described in paragraph \ref{certainty_calculation}. By comparing these two calculation methods, it was possible to evaluate whether the active learning benefited from sampling the most uncertain instances (when using the minimum method), or from sampling the most uncertain images (when using the average method).

Before we could evaluate the effect of the number of forward passes on the active learning performance, a preliminary experiment was performed to examine which numbers were plausible in terms of consistency of the certainty estimate. This consistency was considered important, because when the estimate is consistently uncertain, there is more chance that the image actually contributes to the active learning performance. The setup and results of the preliminary experiment are described in Appendix \ref{appendix1}. Based on the results in Appendix \ref{appendix1}, two numbers of forward passes were chosen: 20 and 40. 

The three dropout probabilities, two certainty calculation methods, and two numbers of forward passes were combined into 12 unique combinations of certainty calculation parameters. We assessed the effect of each of these 12 combinations on the active learning performance. All combinations were tested with the same initial dataset of 100 images, which were randomly sampled from the training pool. After training Mask R-CNN on the initial dataset, the trained model was used to select 200 images from the remaining training pool about which Mask R-CNN was most uncertain. The selected images were used together with the initial training images to retrain Mask R-CNN. This procedure was repeated 12 times, such that in total 13 image sets were trained (containing respectively, 100, 300, 500, ..., 2500 sampled images). After training Mask R-CNN on each image set, the performance of the trained model was determined on the images of the first test set. The 13 resulting mAP values were stored. The experiment was repeated five times to account for the randomised initial dataset and the randomness in the Monte-Carlo dropout method.

A three-way analysis of variance (ANOVA) with a significance level of 5\% was employed for the mAP values to test whether there were significant performance differences between the three dropout probabilities, the two certainty calculation methods, and the two numbers of forward passes. The ANOVA was performed per mAP value, because the mAP was not independent between the different image sets (for instance, the set of 2500 images contained the 2300 images from the previous sampling iteration). 

\subsubsection{Experiment 2}
\label{experiment2}
The optimal setting for the dropout probability, certainty calculation method, and number of forward passes was obtained from experiment 1 and used in experiments 2 and 3. The objective of experiment 2 was to test the effect of the sample size on the active learning performance. This experiment would reveal the optimal sample size for annotating the images and retraining Mask R-CNN. A smaller sample size would possibly be better for the active learning performance, because Mask R-CNN would then have more chances to retrain on the images it was uncertain about. A larger sample size will reduce the total sampling time.

Four sample sizes were tested: 50, 100, 200, and 400 images. These four sample sizes were considered the most practical in terms of annotation and training time. For all sample sizes, the initial dataset size was 100 images and the maximum number of training images was 2500 images. The number of sampling iterations for the four sample sizes was respectively, 48, 24, 12, and 6. The performances of the trained Mask R-CNN models were determined on the images of the second test set. The experiment was repeated five times to account for the randomised initial dataset and the randomness in the Monte-Carlo dropout method. 

A one-way ANOVA with a significance level of 5\% was employed for the mAP values to test whether there were significant performance differences between the four sample sizes. The ANOVA was performed on the mAP values that shared a common number of training images (the common numbers were 500, 900, 1300, 1700, 2100, and 2500 images). The ANOVA was not performed on the mAP value of the initial dataset, as this value was the same between the four sample sizes (since the training was performed with the same dropout probability). 

\subsubsection{Experiment 3}
\label{experiment3}
The objective of experiment 3 was to compare the performance of the active learning with the performance of the random sampling. The initial dataset size was 100 images, and both sampling methods used the same sample size that was chosen from experiment 2. The performances of the trained Mask R-CNN models were determined on the images of the third test set. The experiment was repeated five times to account for the randomised initial dataset and the randomness in the Monte-Carlo dropout method. A one-way ANOVA with a significance level of 5\% was employed to test whether there were significant performance differences between the active learning and the random sampling. The ANOVA was not performed on the mAP value of the initial dataset, as this value was the same between the two sampling methods (since the training was performed with the same dropout probability).

For comparison, another Mask R-CNN model was trained on the entire training pool (14,000 images). The performance of this model was also evaluated on the images of the third test set. The resulting mAP value was considered as the maximum mAP that could have been reached on our dataset. 
\newpage

\section{Results}
The results are summarised per experiment: experiment 1 (paragraph \ref{results_exp1}), experiment 2 (paragraph \ref{results_exp2}), and experiment 3 (paragraph \ref{results_exp3}). 

\subsection{The effect of the dropout probability, the number of forward passes, and the certainty calculation method on the active learning performance (experiment 1)}
\label{results_exp1}
Table \ref{tab:table_exp1_anova} summarises the effects of the dropout probability, the number of forward passes and the certainty calculation method on the active learning performance (mAP). The results of the ANOVA are summarised by the different letters. The letters are sorted in descending order, meaning that letter "a" significantly outperforms letters "b" and "c" at a significance level of 5\% (p=0.05). The performance means that do not have a letter in common are significantly different. 

The dropout probability had the largest effect on the active learning performance (Table \ref{tab:table_exp1_anova}). For all thirteen sampling iterations, dropout probability 0.25 had a significantly higher mAP than dropout probability 0.75. In five sampling iterations (1, 2, 3, 11, and 12), the dropout probability 0.25 had a significantly higher mAP than the dropout probability 0.50. These results were in line with Figure \ref{fig:certainty_fp} (Appendix \ref{appendix1}), which showed that the dropout probability 0.25 had the most consistent certainty estimate. Dropout probability 0.25 was chosen as the preferred probability in the next experiments.

The number of forward passes had a small effect on the active learning performance. In only three of the thirteen sampling iterations, there was a significant difference between 20 and 40 forward passes (Table \ref{tab:table_exp1_anova}). In sampling iterations 8, 11, and 12, the mAP was significantly higher at 20 forward passes. This result was unexpected, as Figure \ref{fig:certainty_fp} (Appendix \ref{appendix1}) showed that there was a less consistent certainty estimate at 20 forward passes than at 40 forward passes. It should be noted that the mAP differences between 20 and 40 forward passes were relatively small (maximally 1.7 mAP). We decided to choose 20 forward passes in the next experiments.

The certainty calculation method had the smallest effect on the active learning performance. In only one of the thirteen sampling iterations, there was a significant difference between the average method and the minimum method (Table \ref{tab:table_exp1_anova}). In sampling iteration 10, the average method had a significantly higher mAP than the minimum method, but the difference was small (1.3 mAP). The small mAP differences were probably due to the limited number of broccoli instances per image. There were on average two broccoli instances per image, suggesting that the choice of the average or the minimum method probably did not have much influence on the active learning performance. Despite the small mAP differences, we decided to choose the average method in the next experiments.

The decision to continue with the parameters for the dropout probability (0.25), number of forward passes (20) and certainty calculation method (average), meant that five significant interactions between the dropout probability and the certainty calculation method and one significant interaction between the number of forward passes and the certainty calculation method were ignored. These significant interactions were all due to the dropout probability 0.75 (whose performance was found to be insufficient). There were no significant interactions between the dropout probability, the number of forward passes and the certainty calculation method. 

\begin{table}[hbt!]
\caption{Performance means expressed for the three dropout probabilities, the two numbers of forward passes, the two certainty calculation methods, and the thirteen sampling iterations. The results of the ANOVA are summarised by the different letters. The letters are sorted in descending order, meaning that letter "a" significantly outperforms letters "b" and "c" at a significance level of 5\% (p=0.05).}
\begin{center}
\begin{filecontents*}{csv/exp1_anova.csv}
iteration,numimg,0.25,0.5,0.75,space1,20,40,space2,average,minimum
1,100,21.5 a,18.4 b,12.4 c,,17.4 a,17.5 a,,17.2 a,17.7 a
2,300,35.8 a,31.3 b,17.5 c,,28.6 a,27.8 a,,27.7 a,28.8 a
3,500,42.1 a,37.5 b,22.0 c,,34.1 a,33.6 a,,33.5 a,34.3 a
4,700,47.9 a,46.1 a,31.5 b,,41.7 a,41.9 a,,41.4 a,42.2 a
5,900,49.8 a,49.6 a,36.3 b,,45.3 a,45.1 a,,45.0 a,45.5 a
6,1100,53.1 a,52.7 a,41.0 b,,49.3 a,48.6 a,,48.9 a,49.0 a
7,1300,54.5 a,53.2 a,46.0 b,,51.7 a,50.8 a,,51.5 a,51.0 a
8,1500,56.5 a,55.3 a,49.7 b,,54.7 a,53.0 b,,53.8 a,53.9 a
9,1700,57.2 a,57.5 a,54.0 b,,56.8 a,55.7 a,,56.8 a,55.6 a
10,1900,59.0 a,58.2 a,55.5 b,,58.0 a,57.2 a,,58.2 a,56.9 b
11,2100,60.1 a,58.4 b,56.9 c,,59.1 a,57.8 b,,58.6 a,58.3 a
12,2300,60.6 a,58.7 b,57.8 b,,59.7 a,58.4 b,,59.2 a,58.9 a
13,2500,61.1 a,60.2 a,58.6 b,,60.4 a,59.5 a,,60.2 a,59.7 a
\end{filecontents*}
\csvreader[tabular = {p{1.3cm}>{\centering}p{1.7cm}>{\centering}p{1.25cm}>{\centering}p{1.25cm}>{\centering}p{1.25cm}>{\centering}p{0.05cm}>{\centering}p{1.25cm}>{\centering}p{1.25cm}>{\centering}p{0.05cm}>{\centering}p{1.25cm}>{\centering\arraybackslash}p{1.25cm}},
    table head=\hline & Number of & \multicolumn{9}{c}{Performance (mAP)} \\\cline{3-11}
    Sampling & training & \multicolumn{3}{c}{Dropout probability} & & \multicolumn{2}{c}{Forward passes} & & \multicolumn{2}{c}{Certainty method} \\
    iteration & images & 0.25 & 0.50 & 0.75 & & 20 & 40 & & average & minimum \\\hline,
    late after line=\\,
    late after last line={\\\hline}]
    {csv/exp1_anova.csv}{iteration=\iter, numimg=\numimg, 0.25=\dofirst, 0.5=\dosecond, 0.75=\dothird, space1=\sfirst, 20=\fpfirst, 40=\fpsecond, space2=\ssecond, average=\avg, minimum=\min}
    {\iter & \numimg & \dofirst & \dosecond & \dothird & \sfirst & \fpfirst & \fpsecond & \ssecond & \avg & \min}%
\end{center}
\label{tab:table_exp1_anova}
\end{table}
\newpage

\subsection{The effect of the sample size on the active learning performance (experiment 2)}
\label{results_exp2}
Table \ref{tab:table_exp2_anova} summarises the effect of the sample size on the active learning performance (mAP). The results of the ANOVA are summarised by the different letters. The performance means that do not have a letter in common are significantly different at a significance level of 5\% (p=0.05) (letter "a" significantly outperforms letters "b" and "c"). The ANOVA was not performed on the mAP value of the initial dataset (100 images). 

For all sampling iterations, sample size 200 had the significantly highest mAP. In five of the six sampling iterations (3 to 7), sample size 200 had a significantly higher mAP than sample size 50. In three of the six sampling iterations (4, 5, and 6), sample size 200 had a significantly higher mAP than sample size 100. There was one significant difference between sample size 200 and 400 (at iteration 2), indicating that sample size 400 had the significantly highest mAP in five of the six iterations.

\begin{figure}[hbt!]
  \centering
  \subfloat[] {\includegraphics[width=0.5\textwidth]{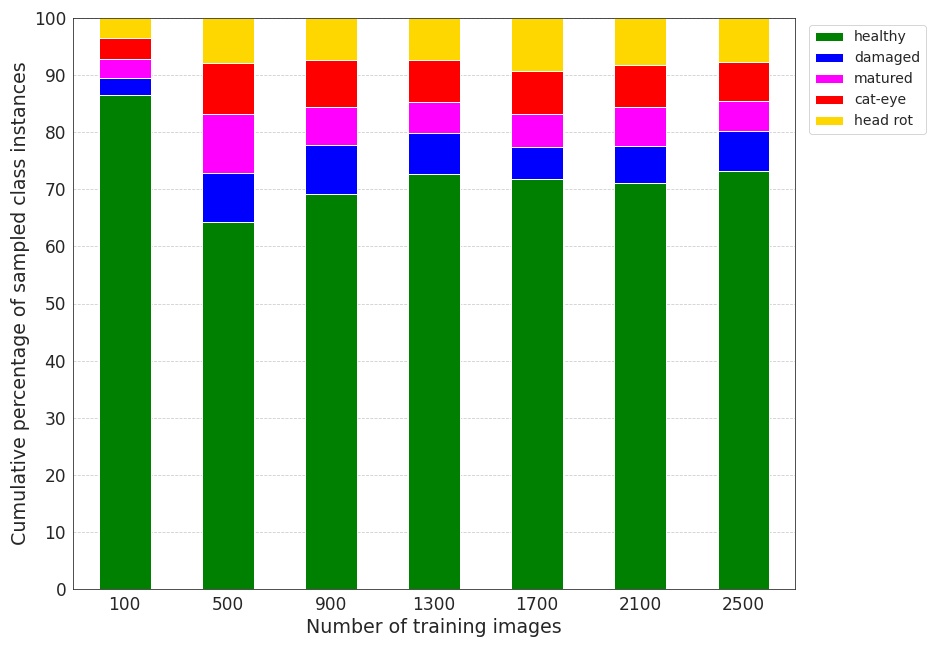}\label{fig:sf48}}
  \hfill
  \subfloat[] {\includegraphics[width=0.5\textwidth]{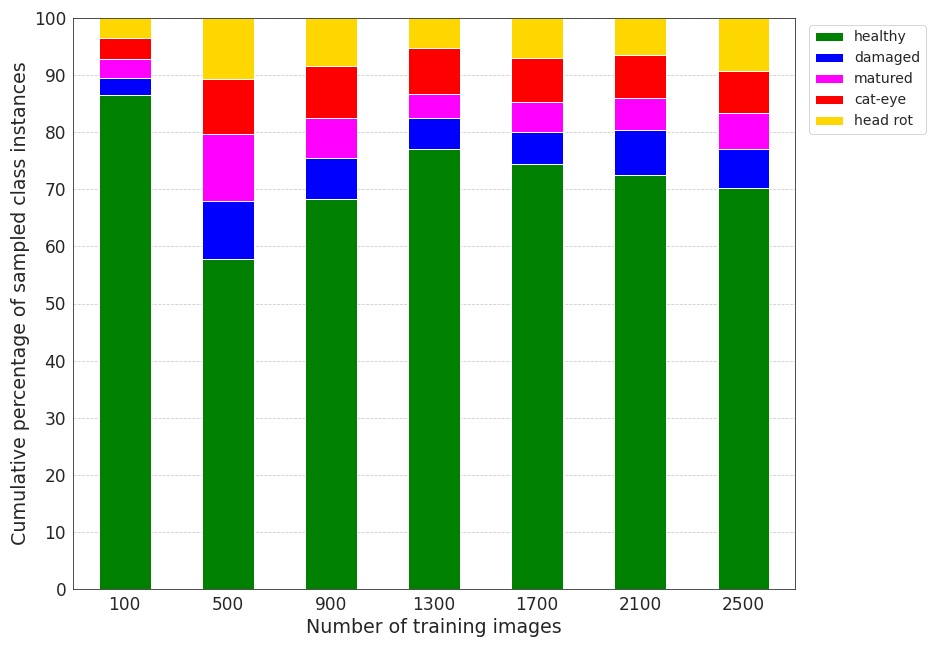}\label{fig:sf24}}
  \hfill
  \subfloat[] {\includegraphics[width=0.5\textwidth]{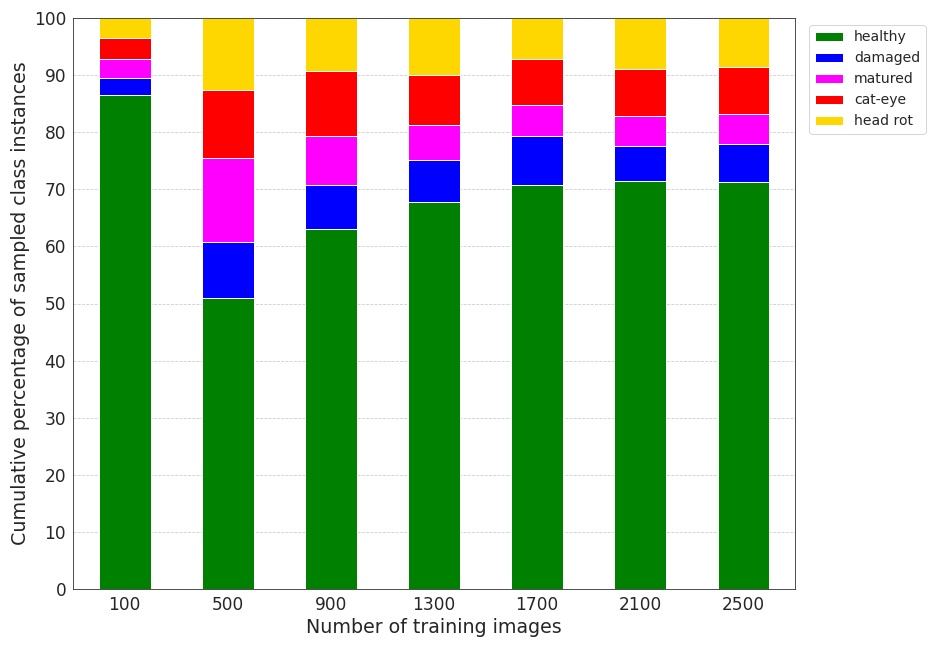}\label{fig:sf12}}
  \hfill
  \subfloat[] {\includegraphics[width=0.5\textwidth]{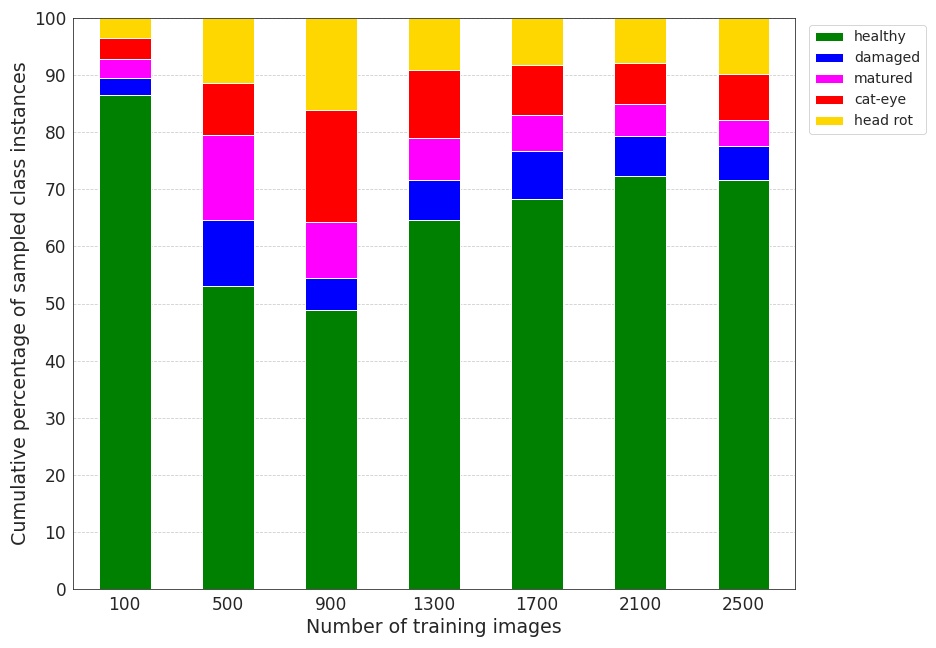}\label{fig:sf6}}
  \hfill
  \caption{Cumulative percentages of the sampled classes in experiment 2. The percentages are expressed for the seven numbers of training images that were shared between the four sample sizes: (a) sample size 50 (b) sample size 100 (c) sample size 200 (d) sample size 400.}
  \label{fig:exp2_sf}
\end{figure}

An in-depth analysis showed that the performance gains of sample sizes 200 and 400 were mainly due to a higher performance on the minority classes cat-eye and head rot. Figures \ref{fig:sf12} and \ref{fig:sf6} show that sample sizes 200 and 400 sampled a higher percentage of these classes compared to sample sizes 50 and 100 (Figures  \ref{fig:sf48} and \ref{fig:sf24}). Apparently, for the active learning performance, it was better to sample with larger image batches that primarily consisted of one or two previously underperforming minority classes, than to sample with smaller image batches that had a more balanced ratio of the minority classes.  

One research demarcation may have influenced the results. In experiment 1, only sample size 200 was tested, and this may have resulted that the chosen parameters from experiment 1 were only optimised for sample size 200 and not for sample sizes 50, 100 and 400. Despite this research demarcation, we decided to choose sample size 200 in the next experiment. 

\begin{table}[hbt!]
\caption{Performance means for the four sample sizes and the seven sampling iterations that shared a common number of training images. The results of the ANOVA are summarised by the different letters. The letters are sorted in descending order, meaning that letter "a" significantly outperforms letters "b" and "c" at a significance level of 5\% (p=0.05). The performance means that do not have a letter in common are significantly different. The ANOVA was not performed on the mAP value of the initial dataset (100 images).}
\begin{center}
    \begin{filecontents*}{csv/exp2_anova.csv}
iteration,numimg,bs50,bs100,bs200,bs400,pval
1,100,22.4 -,22.4 -,22.4 -,22.4 -,-
2,500,40.9 a,41.6 a,42.2 a,36.7 b,0.01 (*)
3,900,47.5 b,48.8 ab,50.5 a,48.4 ab,0.12 (ns)
4,1300,51.0 b,50.3 b,55.2 a,55.6 a,0.00 (****)
5,1700,53.1 b,51.7 b,57.9 a,57.5 a,0.00 (****)
6,2100,54.3 b,56.1 b,58.9 a,60.2 a,0.00 (****)
7,2500,56.7 c,57.3 bc,59.5 ab,60.5 a,0.01 (**)
\end{filecontents*}
\csvreader[tabular = {p{1.7cm}>{\centering}p{2.4cm}>{\centering}p{1.8cm}>{\centering}p{1.8cm}>{\centering}p{1.8cm}>{\centering\arraybackslash}p{1.8cm}},
    table head=\hline & &\multicolumn{4}{c}{Performance (mAP)} \\\cline{3-6}
    Sampling & Number of &\multicolumn{4}{c}{Sample size} \\
    iteration & training images & 50 & 100 & 200 & 400 \\\hline,
    late after line=\\,
    late after last line={\\\hline}]
    {csv/exp2_anova.csv}{iteration=\iter, numimg=\numimg, bs50=\first, bs100=\second, bs200=\third, bs400=\fourth}
    {\iter & \numimg & \first & \second & \third & \fourth}%
\end{center}
\label{tab:table_exp2_anova}
\end{table}

\clearpage

\subsection{Performance comparison between active learning and random sampling (experiment 3)}
\label{results_exp3}
\begin{figure}[hbt!]
  \centering
    \includegraphics[width=1\textwidth]{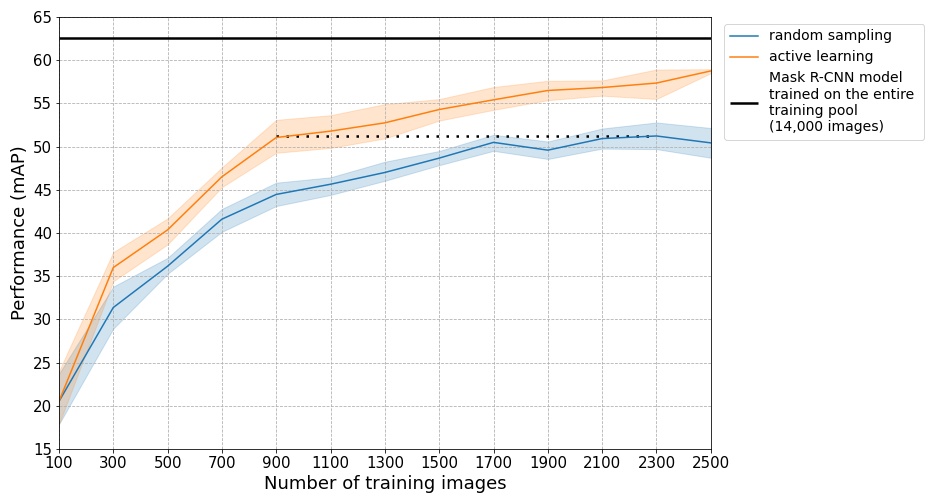}
    \caption{Performance means of the active learning (orange line) and the random sampling (blue line). The coloured areas around the lines represent the 95\% confidence intervals around the means. For all sampling iterations, the active learning had a significantly higher mAP than the random sampling (the ANOVA was not performed on the mAP value of the initial dataset (100 images)). The black solid line represents the performance of the Mask R-CNN model that was trained on the entire training pool (14,000 images). The black dashed line is an extrapolation of the maximum performance of the random sampling to the performance curve of the active learning. The dashed line can be interpreted as the number of annotated images that could have been saved by the active learning while maintaining the maximum performance of the random sampling.}
    \label{fig:exp3}
\end{figure}

\begin{figure}[hbt!]
  \centering
  \subfloat[] {\includegraphics[width=0.5\textwidth]{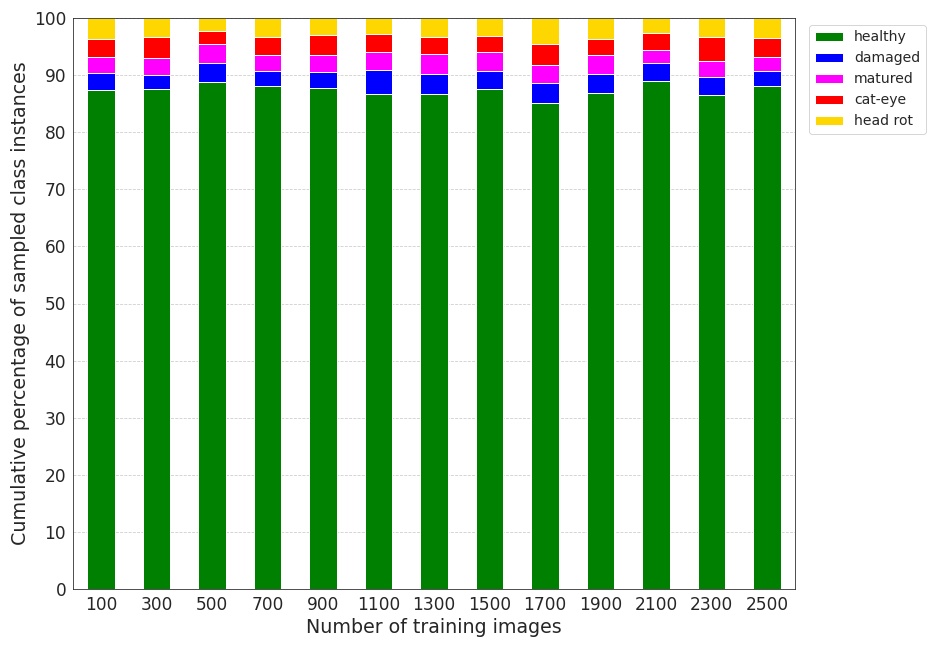}\label{fig:classes_random}}
  \hfill
  \subfloat[] {\includegraphics[width=0.5\textwidth]{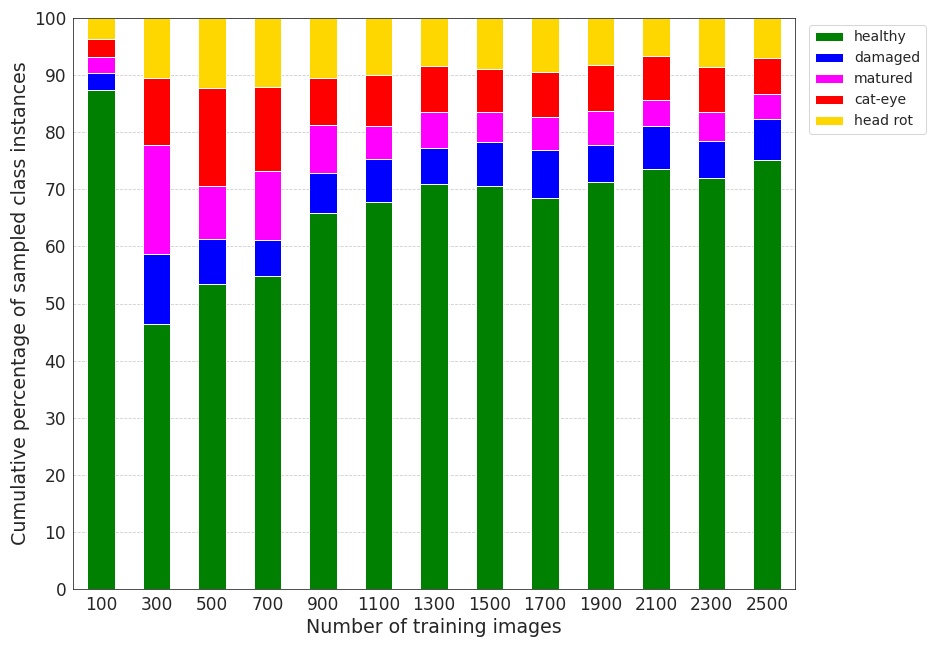}\label{fig:classes_MaskAL}}
  \hfill
  \caption{Cumulative percentages of the sampled classes in experiment 3. The percentages are expressed for the thirteen numbers of training images and the two sampling methods: (a) random sampling (b) active learning.}
  \label{fig:exp3_classes}
\end{figure}

\begin{figure}[hbt!]
  \centering
  \subfloat[] {\includegraphics[width=0.495\textwidth]{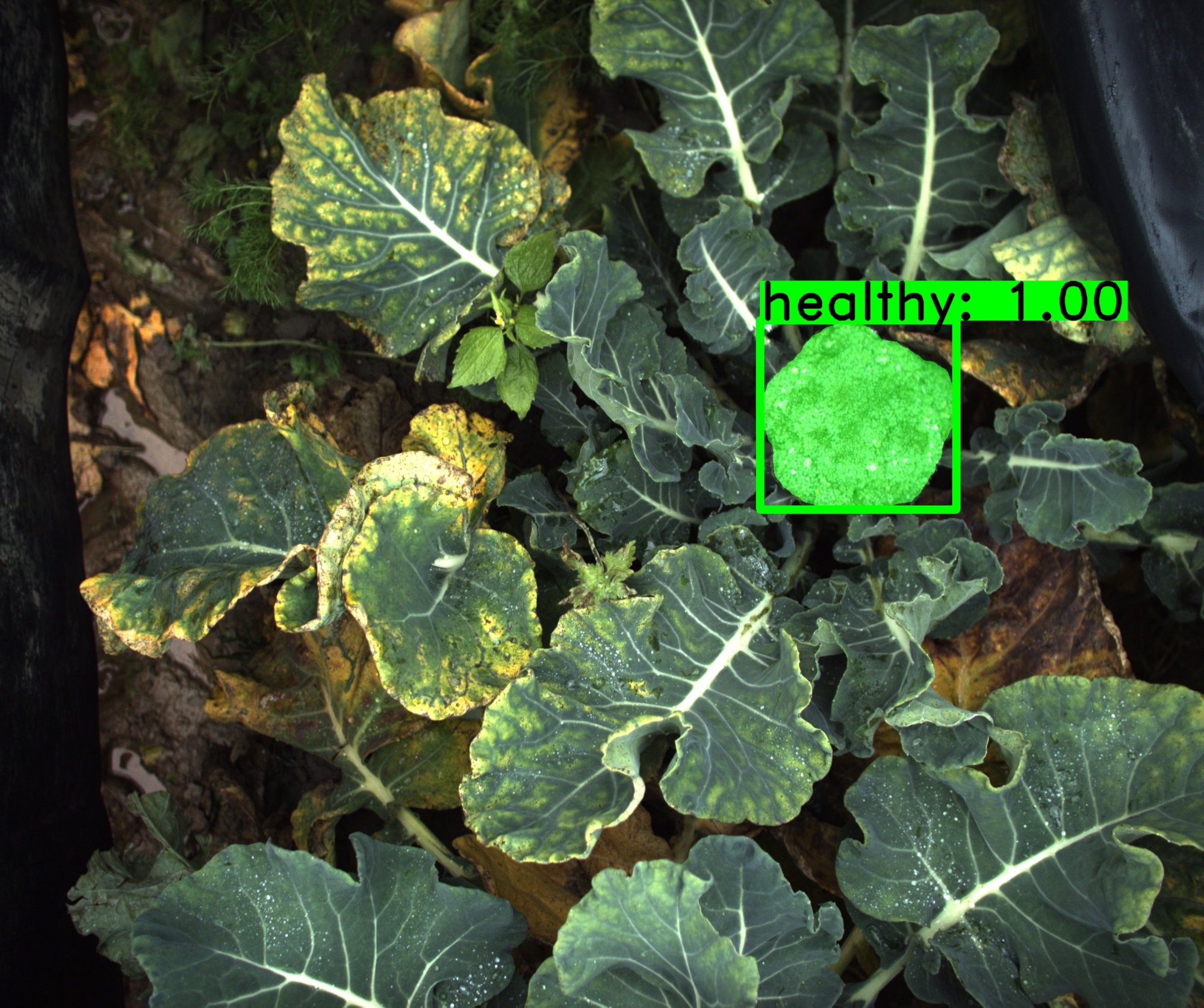}\label{fig:output1_random}}
  \hfill
  \subfloat[] {\includegraphics[width=0.495\textwidth]{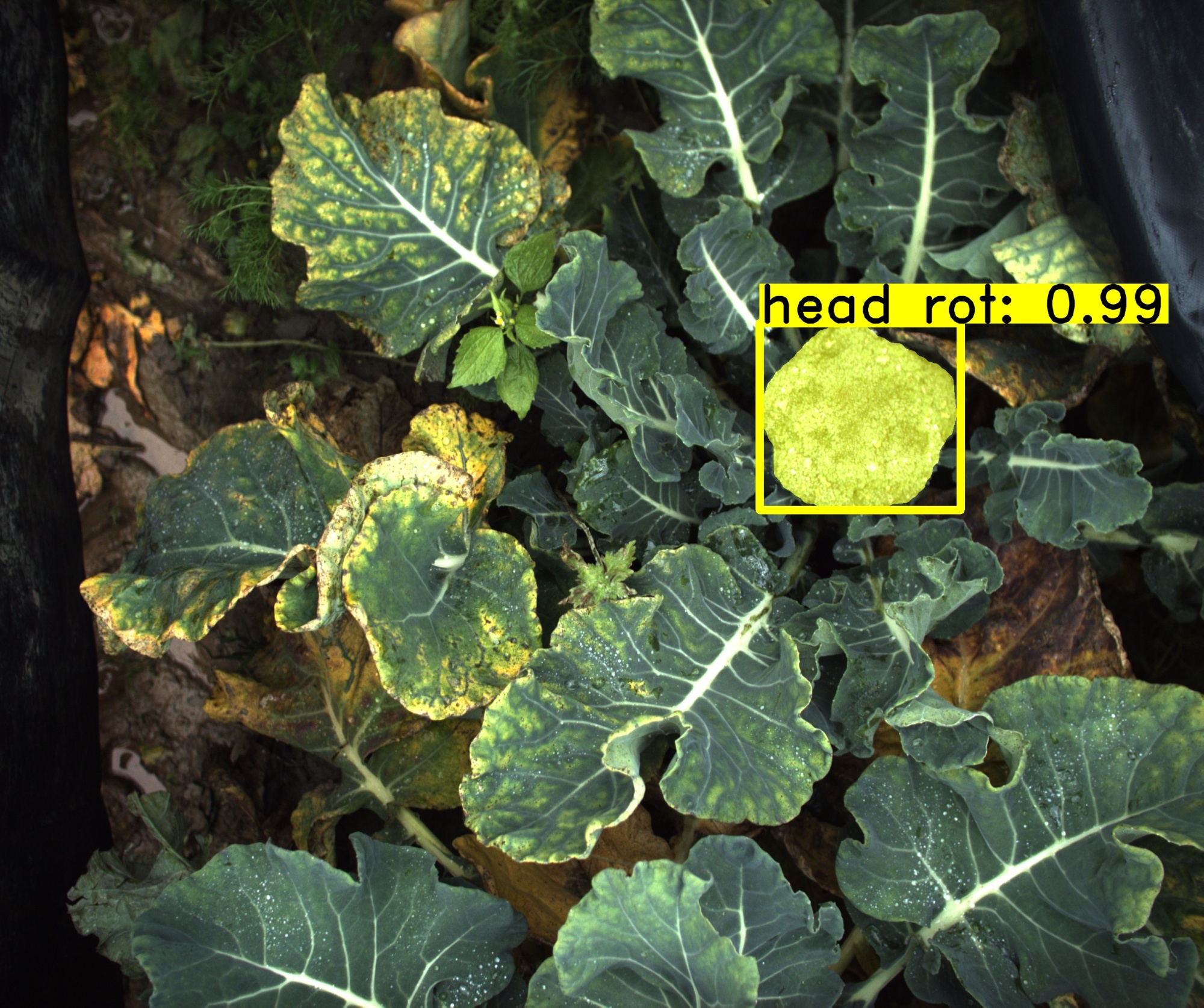}\label{fig:output1_activelearning}}
  \hfill
  \caption{Instance segmentation outputs of Mask R-CNN on the same image with a head rot infected broccoli head. (a) The Mask R-CNN model that was trained with the random sampling method misclassified the broccoli head as being healthy. (b) The Mask R-CNN model that was trained with MaskAL correctly classified the broccoli head.}
  \label{fig:output1}
\end{figure}

\begin{figure}[hbt!]
  \centering
  \subfloat[] {\includegraphics[width=0.495\textwidth]{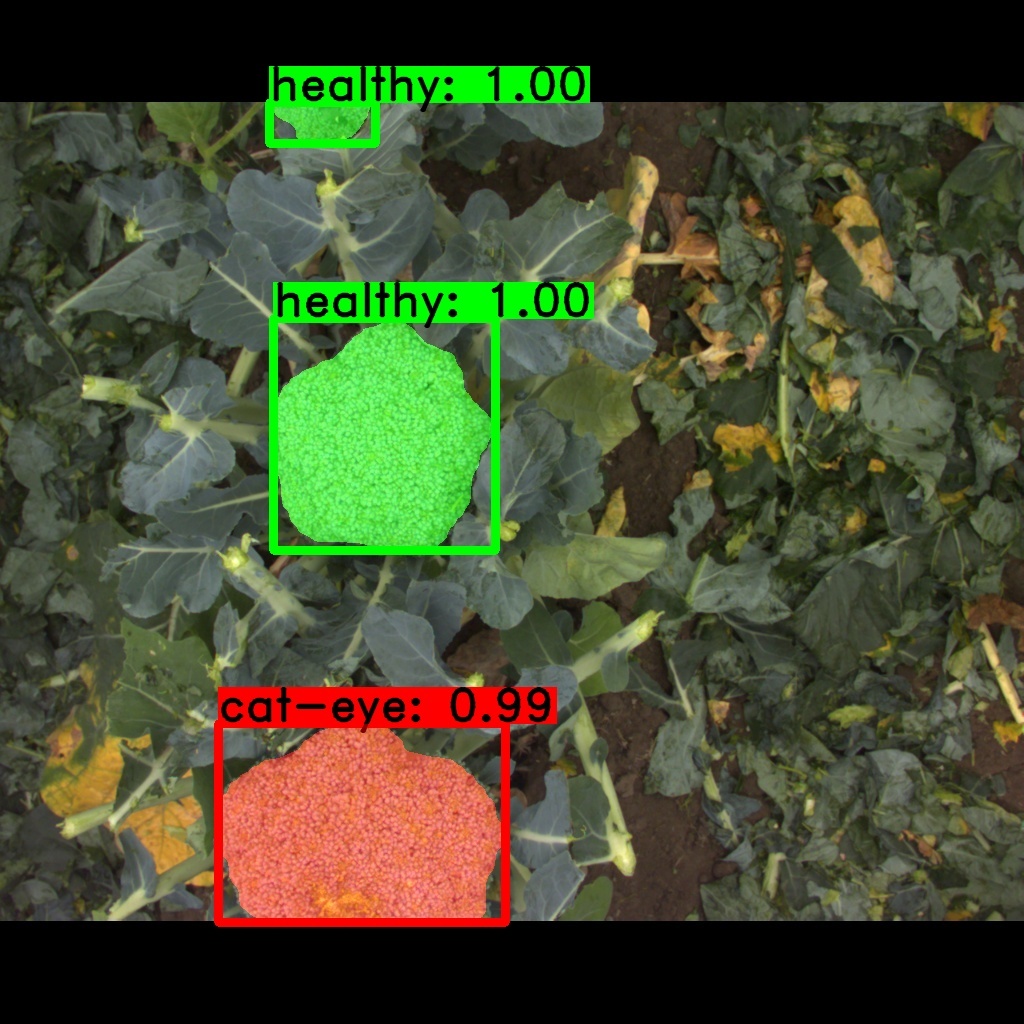}\label{fig:output2_random}}
  \hfill
  \subfloat[] {\includegraphics[width=0.495\textwidth]{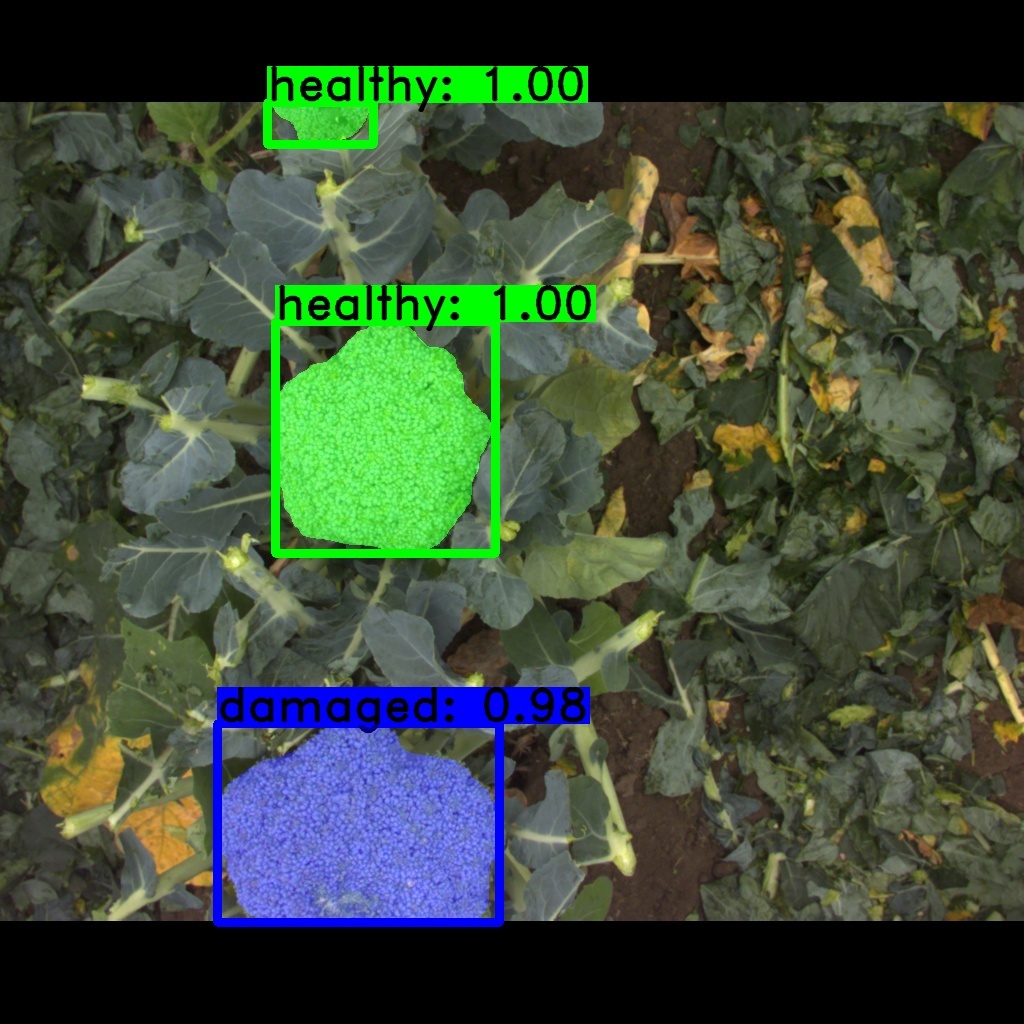}\label{fig:output2_activelearning}}
  \hfill
  \caption{Instance segmentation outputs of Mask R-CNN on the same image with two healthy broccoli heads (the ones in the top and centre of the image) and one damaged broccoli head (in the bottom of the image). (a) The Mask R-CNN model that was trained with the random sampling method misclassified the damaged broccoli head as being cat-eye. (b) The Mask R-CNN model that was trained with MaskAL correctly classified the three broccoli heads.}
  \label{fig:output2}
\end{figure}

Figure \ref{fig:exp3} visualises the performance of the active learning and the random sampling for the thirteen sampling iterations (100, 300, ..., 2500 sampled images). The coloured areas around the lines represent the 95\% confidence intervals around the means. For all sampling iterations, the active learning had a significantly higher mAP than the random sampling (the ANOVA was not performed on the mAP value of the initial dataset (100 images)). The performance differences were between 4.2 and 8.3 mAP. Figure \ref{fig:exp3_classes} visualises the possible cause of the performance differences. With the random sampling, there was a lower percentage of sampled instances of the four minority classes. As a result, the classification performance on these minority classes was lower. The active learning sampled a higher percentage of images with minority classes, leading to a significantly better performance. Figures \ref{fig:output1} and \ref{fig:output2} visualise the Mask R-CNN performance on two broccoli images. 

With the random sampling, the maximum performance was 51.2 mAP and this value was achieved after sampling 2300 images. With the active learning, a similar performance was achieved after sampling 900 images (51.0 mAP), indicating that potentially 1400 annotations could have been saved (see the black dashed line in Figure \ref{fig:exp3}). 

The maximum performance of the active learning was 58.7 mAP and this value was achieved after sampling 2500 images. This maximum performance was 3.8 mAP lower than the performance of the Mask R-CNN model that was trained on the entire training pool of 14,000 images (62.5 mAP). This means that the active learning achieved 93.9\% of that model's performance with 17.9\% of its training data. The maximum performance of the random sampling was 11.3 mAP lower than the performance of the Mask R-CNN model trained on the entire training pool. The random sampling achieved 81.9\% of that model's performance with 16.4\% of its training data.   

Both sampling methods achieved their largest performance gains during the first 1200 sampled images, see Figure \ref{fig:exp3}. The gains were respectively, 32.2 mAP with the active learning and 26.5 mAP with the random sampling. During the last 1200 sampled images, there was a marginal performance increase of 6.0 mAP with the active learning and 4.2 mAP with the random sampling. Moreover, with the random sampling, there was only an increase of 0.7 mAP during the last 800 sampled images. This suggests that the annotation of these 800 images probably would have cost more than it would have benefited.

\newpage

{\clearpage}
\section{Discussion}
\label{discussion}
By using active learning, the performance of Mask R-CNN improved faster, and thereby the annotation effort could be reduced compared to a random sampling method. Although this outcome was only demonstrated on one dataset, our results suggest that better performance can be achieved when retraining Mask R-CNN on images about which the model was most uncertain. 

On our class imbalanced dataset, the better performance of the active learning was due to the sampling of a higher fraction of images containing the four minority classes. It was expected that the sampled images had a low semantic certainty due to the difficulty in correctly classifying the class labels.  On the other hand, it was probably easier for Mask R-CNN to learn the masks of the five broccoli classes, since they had a similar shape. As such, the spatial certainty might have been higher. There was probably also a higher occurrence certainty, as the broccoli heads were generally well visible in the image. Thus, in our dataset, the overall image certainty was probably more influenced by the semantic certainty than by spatial or occurrence certainty. This outcome was opposite to \citet{lopezgomez2019}, who found that the spatial certainty contributed most to the active learning performance. The difference is that López Gómez tested the active learning on a dataset with bicycles and motorcycles, and the shapes of these classes were probably more difficult for Mask R-CNN to learn. Thus, depending on the specific variation and challenges in a dataset, the active learning can emphasise the optimisation of a specific certainty. Since MaskAL is the first active learning framework with three certainty calculations, we expect that it can also be used on datasets with instances with a difficult-to-learn shape or datasets with small and unclear instances. 

There are several ways in which MaskAL could be further improved. First, weighting factors could be integrated into the certainty equation to tune the relative importance of either semantic, spatial, or occurrence certainty. We believe that this could have improved the MaskAL performance on our broccoli dataset, as prioritisation of semantic certainty could have led to faster optimisation of the classification performance. Second, the dropout can be further optimised. As demonstrated in experiment 1, the choice of the dropout probability had a large influence on the active learning performance. To choose the optimal dropout probability on a different dataset, it may be necessary to redo experiment 1, possibly with more than three tested probabilities. However, it is time consuming to conduct such an experiment for every new dataset. Therefore, we recommend investigating whether the concrete dropout method of \citet{gal2017} can be used as an alternative method to automatically optimise the dropout probability during training. After using the concrete dropout method, the optimised dropout probability can be used for sampling, and this might reduce the experimental time and improve the active learning performance. 

Although our active learning method and dataset were different, it is possible to qualitatively compare our results with those of \citet{lopezgomez2019}, \citet{dijk2019}, and \citet{wang2020}. In the study of \citet{lopezgomez2019}, the active learning with Monte-Carlo dropout performed better than the random sampling in five of the eight sampling iterations. In the other three sampling iterations, the active learning performed worse than the random sampling. One possible reason for this poorer performance was that \citet{lopezgomez2019} sampled both images about which Mask R-CNN was uncertain as images about which Mask R-CNN was certain (this was done to increase the diversity in the image set). We believe that \citet{lopezgomez2019} could have achieved better results if Mask R-CNN had been retrained only on the images about which the model was most uncertain. In the study of \citet{dijk2019}, the probabilistic active learning (PAL) did not perform better than the random sampling, possibly because PAL was designed to provide certainty scores only for classification and not for mask segmentation. Another reason for this outcome was that \citet{dijk2019} performed the image sampling on such small datasets that the added value of the active learning may have been limited (the datasets contained respectively 61 images and 45 images). In the study of \citet{wang2020}, the active learning with a learning loss method was compared to random sampling on two medical datasets. On one dataset, the active learning performed better than the random sampling in the first three sampling iterations, while in the last two iterations the performance was equal. On the other dataset, the active learning performed better than the random sampling in all five sampling iterations. Thus, learning loss can be another promising method for active learning, especially because, unlike MaskAL, it can predict the image uncertainty in one forward pass. In future research, MaskAL should be compared quantitatively to other active learning methods for Mask R-CNN. We also recommend comparing MaskAL's uncertainty sampling with sampling methods other than random sampling, such as diversity sampling or hybrid sampling. Such a comparison could give a better impression of how MaskAL would compare to other more advanced sampling methods. We recommend performing such a comparison on a benchmark dataset, such as Microsoft COCO, because this dataset contains more images and more variation than our broccoli dataset.

The potential use of MaskAL is greater than sampling images from a fixed dataset in an offline setting. MaskAL can also be applied to an operational robot to immediately select the images about which Mask R-CNN is most uncertain. This image selection can be done with a fixed threshold on the image certainty value. The down side of using MaskAL during robot deployment is that the image analysis will take more time. Should the image analysis take more time than desired, then it is recommended to temporarily store the images on the computer, so that they can be analysed by MaskAL after the robot has completed its task. 

MaskAL's certainty calculation can also be used for purposes other than active learning. For instance, the certainty values can be used as an input for the robot to make more targeted decisions, like transporting the harvested broccoli heads with low semantic certainty to another bin for further inspection. The certainty values of MaskAL could also be fused with other predictions or sensor measurements in a probabilistic framework, allowing the robot to better reason under uncertainty. Future research should focus on applying MaskAL for such purposes.
\clearpage

\section{Conclusions}
\label{conclusions}
On our broccoli dataset with five visually similar classes, the active learning with MaskAL performed significantly better than the random sampling. Furthermore, MaskAL had the same performance after sampling 900 images as the random sampling had after sampling 2300 images. This means that by using MaskAL, 1400 annotations could have been saved. Compared to a Mask R-CNN model that was trained on the entire training set (14,000 images), MaskAL achieved 93.9\% of that model's performance with 17.9\% of its training data. In comparison, the random sampling achieved 81.9\% of that model's performance with 16.4\% of its training data. We conclude that by using MaskAL, the annotation effort can be reduced for training Mask R-CNN on a broccoli dataset with visually similar classes. 

In this paper, MaskAL was used for active learning with the purpose of reducing annotation effort. The research was performed on a dataset in which all classes were known. We think that MaskAL can also be valuable for selecting unknown classes in open-set learning. Furthermore, MaskAL can also be used as an uncertainty estimator in probabilistic robotic frameworks. Our software is available for such purposes. 

{\clearpage}
\section*{CRediT authorship contribution statement}
Pieter M. Blok: conceptualisation, methodology, software, data curation, writing - original draft; Gert Kootstra: supervision, conceptualisation, writing - review \& editing; Hakim Elchaoui Elghor: validation, funding acquisition, writing - review \& editing; Boubacar Diallo: conceptualisation, validation, funding acquisition; Frits K. van Evert: supervision, conceptualisation, writing - review \& editing; Eldert J. van Henten: supervision, conceptualisation, writing - review \& editing.

\section*{Funding}
This research was funded by: Topsector TKI AgroFood under grant agreement LWV19178 for “PPP Handsfree production in agri-food”, Agrifac Machinery B.V. and Exxact Robotics.

\section*{Acknowledgements}
We would like to thank Jean-Marie Michielsen and Hyejeong Kim for the image annotation and Paul Goedhart for the statistical analysis. Furthermore, we appreciate the brainstorm sessions we had with our colleagues Manya Afonso, Ard Nieuwenhuizen, Janne Kool, Keiji Jindo, and Andries van der Meer. Finally, we would like to thank Rik van Bruggen from Agrifac Machinery B.V. and Colin Chaballier from Exxact Robotics for their support.  

{\clearpage}

% be aware that the references.bib were altered to display the DOI (by changing the "doi" field to a "note" field)
\bibliographystyle{apalike}
\bibliography{references}
\clearpage

\appendix
\renewcommand\thefigure{\thesection.\arabic{figure}}
\renewcommand\theequation{\thesection.\arabic{equation}}
\setcounter{figure}{0}
\setcounter{equation}{0} 
\section{Appendix A - Preliminary experiment on the effect of the number of forward passes on the consistency of the certainty estimate}
\label{appendix1}
This section summarises the setup and the results of the preliminary experiment that was done to test the effect of the number of forward passes on the consistency of the certainty estimate. This preliminary experiment was conducted with a Mask R-CNN model that was trained on the entire training pool (14,000 images). 

Eighteen numbers of forward passes (\textit{fp}) were tested: from 2 to 10 in steps of 1 and from 10 to 100 in steps of 10. For each number of forward pass, the certainty value was calculated on each instance set that was predicted on the images of the first test set. Then, the absolute difference was calculated between that certainty value, $\texttt{c}_{\texttt{h}_{fp}}$, and the certainty value at 100 forward passes, $\texttt{c}_{\texttt{h}_{100}}$, see Equation \ref{eq_deltauh}. 100 forward passes was the maximum number that could be performed with our graphical processing unit. The $\texttt{c}_{\texttt{h}_{100}}$ value was assumed to approximate the certainty value after an infinite number of forward passes. By calculating the absolute difference with the $\texttt{c}_{\texttt{h}_{100}}$ value, it was possible to get an indication of the consistency of the certainty estimate at a specific forward pass, $\Delta {c}_{\texttt{h}_{fp}}$.

\begin{equation}
\label{eq_deltauh}
\Delta {c}_\texttt{h}(S)_{fp} = |{c}_\texttt{h}(S)_{fp} - {c}_\texttt{h}(S)_{100}| \quad \quad \mbox{with \textit{fp} = $\set{2,3,\ldots,10,20,\ldots,100}$}
\end{equation}

\begin{figure}[hbt!]
  \centering
    \includegraphics[width=0.65\textwidth]{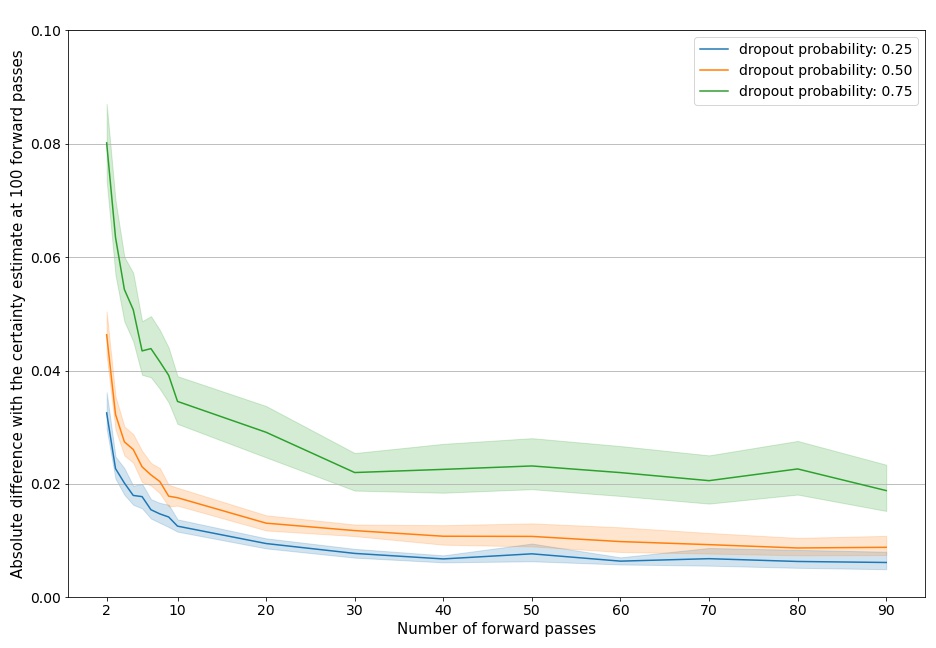}
    \caption{The absolute difference in the certainty estimate between a specific forward pass and 100 forward passes, expressed for the three dropout probabilities. The thick coloured lines are the mean absolute differences and the coloured areas around the lines represent the 95\% confidence intervals around the means.}
    \label{fig:certainty_fp}
\end{figure}

Figure \ref{fig:certainty_fp} visualises the absolute difference in the certainty estimate between a specific forward pass and 100 forward passes. The absolute differences are visualised for the dropout probabilities 0.25, 0.50, and 0.75. For all dropout probabilities, the largest absolute difference was observed for the forward passes lower than 10. Between 10 and 40 forward passes, there was a gradual decrease in the absolute difference. Between 40 and 90 forward passes, the absolute difference was rather constant. Based on Figure \ref{fig:certainty_fp}, we chose 20 and 40 forward passes to be tested in experiment 1. These values produced a relatively consistent certainty estimate (especially for the dropout probabilities 0.25 and 0.50). In addition, the value 20 was closest to the 16 forward passes that were used by \citet{morrison2019}. 

\end{document}